\documentclass[11pt]{article}
\usepackage[UTF8, scheme=plain]{ctex}
\usepackage[final]{acl}

\usepackage{latexsym}

\usepackage{amsmath}
\usepackage{amssymb}

\usepackage{newtxtext}  

\usepackage[T1]{fontenc}
\usepackage[utf8]{inputenc}
\usepackage{microtype}
\usepackage{todonotes}
\usepackage{inconsolata}
\usepackage{array}
\usepackage{graphicx}
\usepackage{subcaption}
\usepackage[table]{xcolor}

\usepackage{multirow}
\usepackage{textalpha}
\usepackage{enumitem} 

\usepackage{booktabs}   
\usepackage{tabularx}
\newcolumntype{L}{>{\raggedright\arraybackslash}X}

\newcolumntype{C}{>{\centering\arraybackslash}X}

\usepackage{algpseudocode}
\usepackage{amsmath}
\usepackage{hyperref}
\usepackage{cleveref}

\usepackage[linesnumbered,ruled,vlined]{algorithm2e}

\usepackage{listings}
\usepackage[most]{tcolorbox}

\newtcolorbox{promptbox}[1]{
    enhanced,
    breakable,
    colframe=black!60,
    colback=white,
    colbacktitle=black!60,
    coltitle=white,
    fonttitle=\bfseries\large,
    title={#1},
    halign title=left,
    arc=2mm,
    boxrule=1.0pt,
    toprule at break=1.0pt,
    bottomrule at break=1.0pt,
    left=6pt, right=6pt, top=6pt, bottom=6pt,
    toptitle=2mm, bottomtitle=2mm
}

\newcommand{\eg}[1]{\textit{e.g.}}
\newcommand{\ie}[1]{\textit{i.e.}}
\newcommand{\etal}[1]{\textit{et al.}}
\newcommand{\etc}[1]{\textit{etc.}}

\SetCommentSty{graycomment}

\usepackage{graphicx}
\usepackage{subcaption}
\usepackage{hyperref}
\usepackage{url}
\usepackage{multirow}
\usepackage{wrapfig}
\usepackage{array}
\usepackage{caption}
\usepackage{arydshln} 
\usepackage{booktabs}
\usepackage{colortbl}
\definecolor{lightgray}{rgb}{0.9,0.9,0.9}
\definecolor{mygray}{gray}{.9} 
\usepackage{amsmath}    
\usepackage{mathrsfs}   
\usepackage{xcolor}  
\usepackage{textcomp}
\usepackage{booktabs}
\usepackage{arydshln}
\usepackage{booktabs}
\definecolor{darkblue}{rgb}{0.0, 0.17, 0.58}
\definecolor{darkgreen}{rgb}{0.0, 0.5, 0.0}

\SetCommentSty{graycomment}

\title{Dr. Assistant: Enhancing Clinical Diagnostic Inquiry via Structured Diagnostic Reasoning Data and Reinforcement Learning}

\usepackage{authblk}

\renewcommand{\thefootnote}{%
  \ifcase\value{footnote}
    \or *    
    \or \dag 
    \or \ddag
    \else *  
  \fi
}

\author[1]{Yue Guo\textsuperscript{*}}
\author[1]{Fanfu Wang}
\author[1]{Jianwei Lv\textsuperscript{*}}
\author[1]{Xincheng Shi}
\author[1]{Yuchen Li}
\author[1]{Yunsheng Zeng} 
\author[1]{Youya \\Wang}
\author[1]{Yujing Liu}
\author[1]{Yunhao Qiao}
\author[1]{Gen Li}
\author[1]{Junfeng Wang\textsuperscript{*}}
\author[1]{Bo Yuan\textsuperscript{*}\textsuperscript{\dag}}

\affil[1]{Baidu Inc.}

\begin{document}
\maketitle
\begin{abstract}
Clinical Decision Support Systems (CDSSs) provide reasoning and inquiry guidance for physicians, yet they face notable challenges, including high maintenance costs and low generalization capability.
Recently, Large Language Models (LLMs) have been widely adopted in healthcare due to their extensive knowledge reserves, retrieval, and communication capabilities. While LLMs show promise and excel at medical benchmarks, their diagnostic reasoning and inquiry skills are constrained.
To mitigate this issue, we propose (1) Clinical Diagnostic Reasoning Data (CDRD) structure to capture abstract clinical reasoning logic, and a pipeline for its construction, and (2) the Dr. Assistant, a clinical diagnostic model equipped with clinical reasoning and inquiry skills. Its training involves a two-stage process: SFT, followed by RL with a tailored reward function.
We also introduce a benchmark to evaluate both diagnostic reasoning and inquiry.
Our experiments demonstrate that the Dr. Assistant outperforms open-source models and achieves competitive performance to closed-source models, providing an effective solution for clinical diagnostic inquiry guidance. Project information can be found at: \href{https://github.com/YGswu/Dr.-Assistant}{https://github.com/YGswu/Dr.-Assistant}.
\end{abstract}

\footnotetext[1]{\footnotesize Project leader: \{guoyue07,lvjianwei,wangjunfeng,\\yuanbo07\}@baidu.com}
\footnotetext[2]{\footnotesize Corresponding author: yuanbo07@baidu.com}

\begin{figure*}[t!]
    \centering
    \includegraphics[width=1\textwidth]{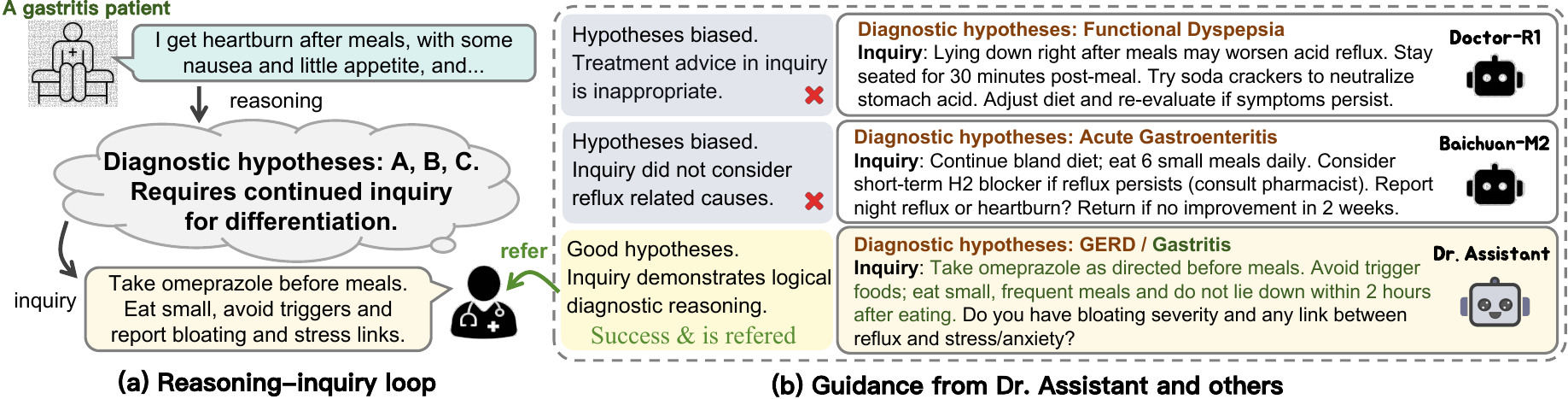} 
    \caption{Dr. Assistant provides guidance (b) to physicians in the reasoning-inquiry loop (a), where diagnostic hypotheses both drive and are refined by ongoing inquiry. Since diagnostic hypotheses are central to reasoning, they are our primary focus. We detail (a) in Appendix \autoref{app:loop}.}
    \label{fig:teaser}
\end{figure*}

\begin{figure}[t!]
    \centering
      \includegraphics[width=0.5\textwidth]{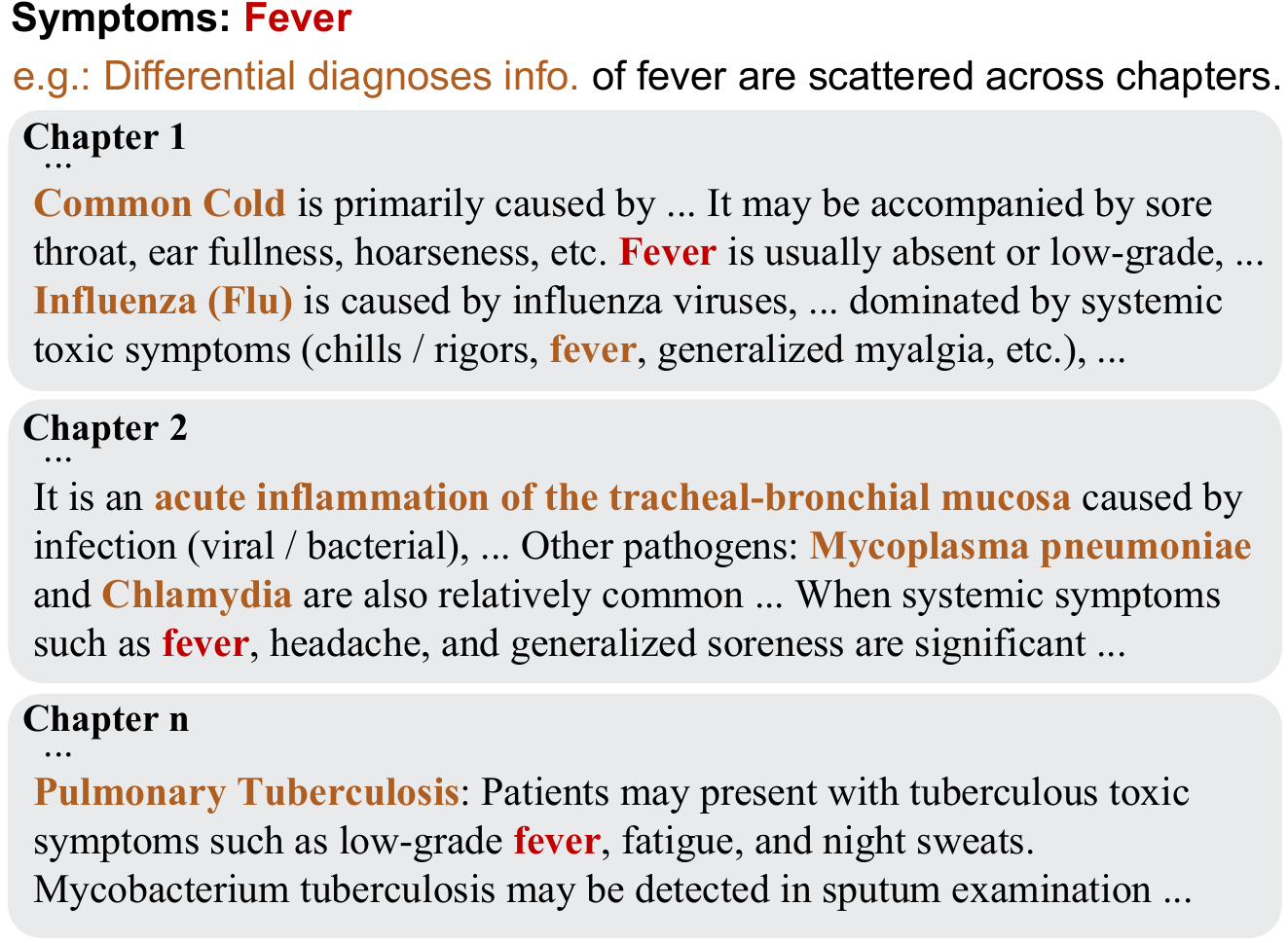} 
    \caption{The scattered logic in clinical guideline.}
    \label{fig:introduce-case}
\end{figure}

\section{Introduction}

Clinical Decision Support Systems (CDSSs) assist physicians by providing guidance for clinical inquiries, which is especially valuable for less experienced physicians \cite{bright2012effect,sutton2020overview,berner2016overview}.
However, traditional CDSSs rely on structured knowledge bases and rule-based algorithms, which incur high development and maintenance costs and have limited adaptability to complex clinical situations \cite{sheikhalishahi2019natural,deo2015machine}.

Recently, Large Language Models (LLMs) have achieved remarkable success in knowledge-intensive tasks, spanning legal document analysis \cite{19,20}, financial forecasting \cite{21,22} and education \cite{23,24}, due to their superior comprehension and generalization abilities. This progress has spurred their adoption in healthcare \cite{2,18}. 
Baichuan-M2 \cite{2} and HuatuoGPT-o1 \cite{huatuogpt-o1} focus on patients' consultation experience and decision-making. 
Zhongjing \cite{25} prioritizes medical knowledge integration and records analysis.
Current LLMs perform well in general healthcare consultation. However, under zero-shot prompting, they have clear limitations in clinical inquiry as shown in \autoref{fig:teaser} (b), as clinical inquiry requires professional diagnostic reasoning logic.

Clinical guidelines contain diagnostic reasoning logic, yet they face two primary limitations.  1) \textbf{Scattered Reasoning Logic}: The diagnostic reasoning logic cue for a given symptom is often scattered across different chapters, as shown in \autoref{fig:introduce-case}, and 2) \textbf{Difficulty in Training}: Even with high-quality data, training models to master clinical inquiry skills remains a significant challenge. 

To address these challenges, we propose \textbf{C}linical \textbf{D}iagnostic \textbf{R}easoning \textbf{D}ata (CDRD) format to capture abstract diagnostic reasoning logic. Each CDRD is a triple-tuple, including a core symptom, diagnostic evidence and differential diagnoses. We also propose a \textit{three-stage} pipeline for constructing CDRDs. The pipeline collaboratively employs LLM synthesis and physician refinement based on clinical guidelines. It first extracts symptoms (Stage I) and their corresponding diseases (Stage II) and finally completes CDRD (Stage III). To ensure reliability, outputs at every stage are reviewed and refined by professional physicians.
Furthermore, using CDRD as a seed, we synthesize two types of data: Question-Answer pairs (QA data) for Supervised Fine-Tuning (SFT), and clinical inquiry dialogue for Reinforcement Learning (RL).

We also propose \textbf{Dr. Assistant}, a clinical diagnostic model equipped with inherent clinical reasoning \& inquiry skills. Its training consists of two stages:
1) The model remembers preliminary clinical inquiry reasoning logic via SFT.
2) We design a reward function, which encompasses two core evaluation dimensions via RL: Clinical Reasoning and Inquiry Skills, and Fidelity to CDRD (logical differences from CDRD).
The logical differences penalty term (Fidelity to CDRD) is incorporated to constrain illogical inference behaviors, refining the model's clinical reasoning competencies.
Furthermore, we construct a benchmark to evaluate clinical diagnostic reasoning and inquiry skills. It comprises 242 real clinical cases covering 8 secondary departments, and 147 real clinical inquiry rounds covering 4 secondary departments.

The experimental results demonstrate that Dr. Assistant (14 B) outperforms the open-source model HuatuoGPT-o1-72B with a 13.59\% improvement in ICD-Recall, and achieves comparable performance to GPT-5, which provides an effective solution for the practical implementation of clinical decision support systems.

In summary, our contributions are as follows:
\begin{itemize}
    \item We propose Clinical Diagnostic Reasoning Data (CDRD) structure to capture abstract clinical reasoning logic, and a pipeline for constructing it.
    \item We propose the Dr. Assistant, a clinical diagnostic model equipped with clinical reasoning and inquiry skills. Its training involves a two-stage process: SFT, followed by RL with a tailored reward function to improve the model's clinical reasoning and inquiry skills.
    \item We construct a clinical diagnostic reasoning \& inquiry benchmark consisting of 242 real cases across 8 secondary departments, along with 147 real inquiry turns across 4 secondary departments. The experiments demonstrate Dr. Assistant outperforms the open-source model HuatuoGPT-o1-72B, with a 13.59\% improvement in ICD-Recall rate, and achieves comparable performance to GPT-5.
\end{itemize}
\begin{figure*}[t!]
    \centering
    \includegraphics[width=\textwidth]{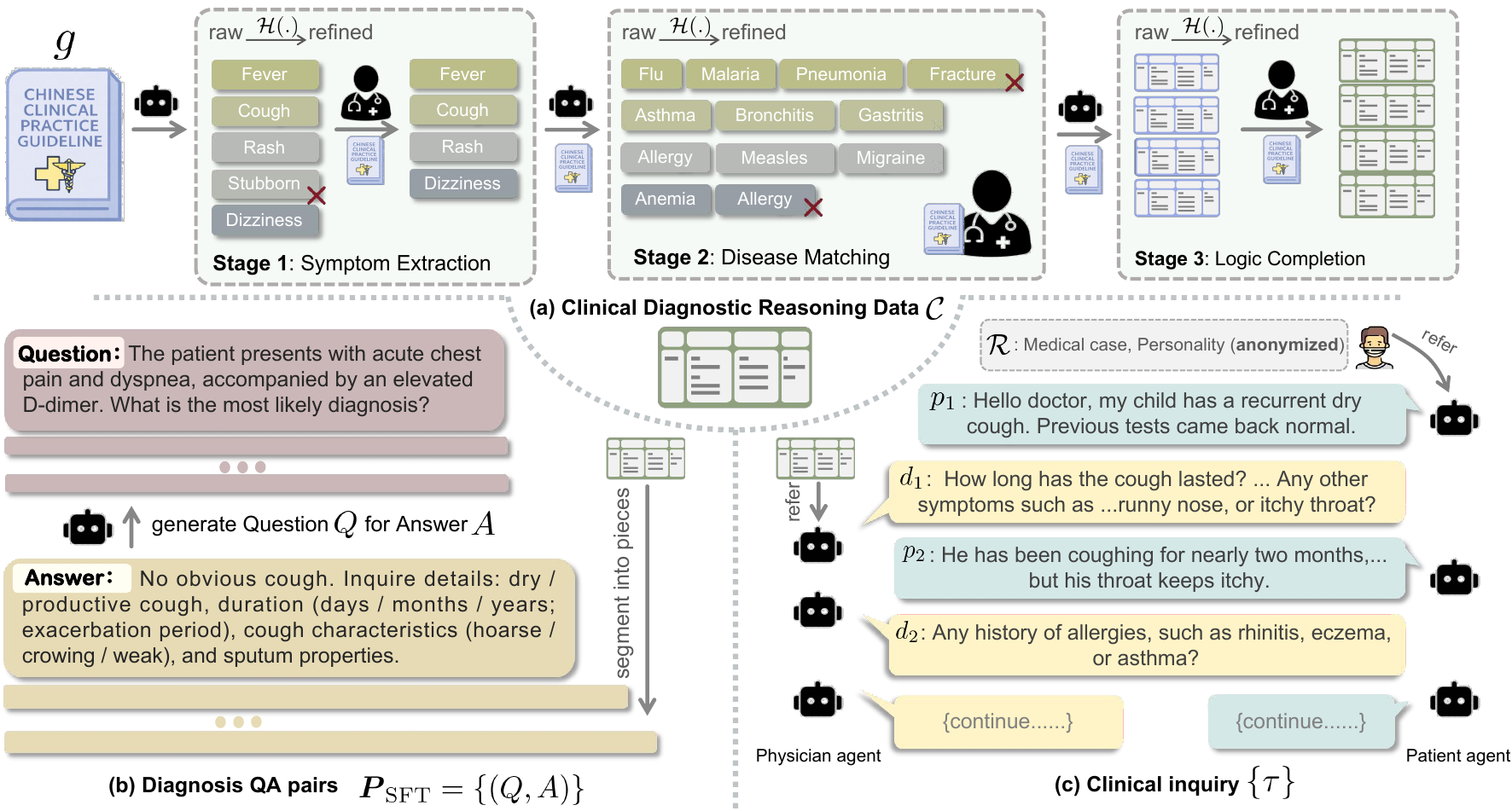}
    \includegraphics[width=\textwidth]{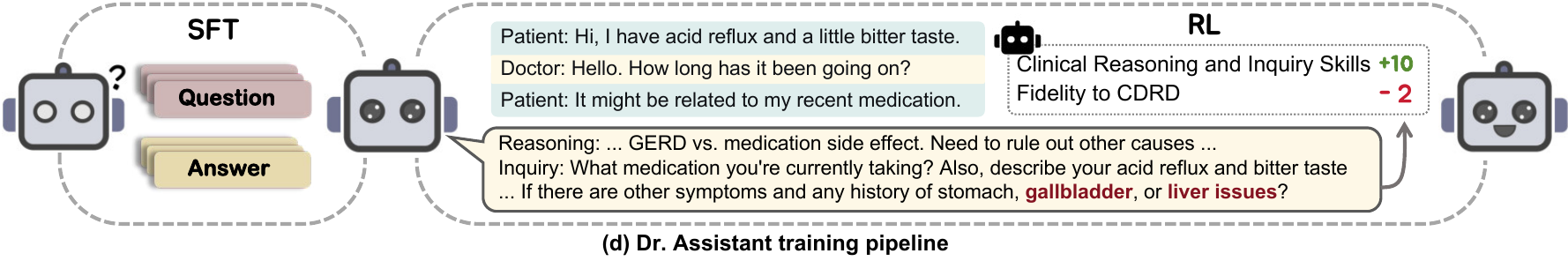}
    \caption{Our workflow involves CDRD construction (a), data synthesis (b, c), and Dr. Assistant training (d).}
    \label{fig:pipe_data}
\end{figure*}


\section{Related Work}

\noindent\textbf{LLMs for medical consultation.}
Previous approaches primarily rely on SFT to improve medical knowledge coverage~\cite{17}.
Recent studies incorporate RL to optimize multi-turn consultation strategies, such as Baichuan-M2~\cite{2}, HuatuoGPT-o1~\cite{huatuogpt-o1} and Doctor-R1~\cite{lai2025doctor}, which focus on improving patients' consultation experience.
They are optimized for general conversation, lacking clinical diagnostic inquiry and reasoning skills. This leads to clinically unsupported guidance.

\noindent\textbf{CDSSs.}
Clinical guidelines served as the foundation of traditional CDSSs \cite{berner2016overview,sutton2020overview}.
While rule-based systems provide strong interpretability, they are difficult to scale and adapt to flexible, multi-turn clinical inquiry.
Recent efforts have explored integrating structured medical knowledge with neural models~\cite{sheikhalishahi2019natural}, yet not diagnostic inquiry logic training.

\noindent\textbf{Evaluation of medical consultation.}
Evaluations for Medical LLMs evolved from static question-answer \cite{jin2021disease,pal2022medmcqa,jin2019pubmedqa} to general consultation, such as HealthBench \cite{arora2025healthbench}, MAQuE \cite{gong2025dialogue}, GAPS \cite{chen2025gaps} and SPTesting \cite{dou2024integrating}.
While existing benchmarks feature multi-turn consultation, they frequently fail to capture the realistic demands of clinical diagnostic inquiry in CDSS settings.
In this work, we evaluate models' reasoning soundness and inquiry quality with real and dynamic clinical inquiry cases.

\section{Data Construction}

\label{sec:cr_cot}

\subsection{CDRD Format for Abstract Logic}
Clinical guidelines contain abstract diagnostic reasoning logic and it is scattered across chapters, requiring semantic understanding and re-structure \cite{zhang2305huatuogpt}.
To capture abstract diagnostic reasoning logic that proceeds from a symptom to its differentials in clinical inquiry, we structure it as Clinical Diagnostic Reasoning Data (CDRD).
Formally, each CDRD is defined as a triple-tuple $\mathcal{C} = (\mathcal{S}, \mathcal{E}, \mathcal{D})$, representing the standard diagnostic logic, where:
\begin{itemize}
    \item $\mathcal{S}$: A Core Symptom (e.g., headache) of patient.
    \item $\mathcal{E}$: Diagnostic Evidence. The clinical information, including associated symptoms, lab results, and exam findings, that physicians collect and analyze to infer diagnoses.
    \item $\mathcal{D}$: Differential Diagnoses, the list of potential diseases (e.g., gastroenteritis) should be considered based on the available evidence, along with their clinical manifestations and required diagnostic examinations.
\end{itemize}
Please refer to Appendix \autoref{app:CDRD-demo} for the CDRD demo.

\subsection{CDRD Construction Pipeline}
To maximize the completeness and purity of CDRD, we propose a pipeline for constructing it from clinical guidelines. As shown in \autoref{fig:pipe_data} (a), CDRD construction pipeline is a collaborative approach of LLM synthesis and physician refinement, encompassing three sequential stages: symptom extraction, disease matching and logic completion.

\noindent\textbf{Symptom extraction.}
Given a clinical guideline $g$, we first extract the symptoms set $\boldsymbol{S}$. The LLM identifies candidate symptoms, which are then refined by physicians:
\begin{align}
    \boldsymbol{S} &= \mathcal{H}\Bigl(\operatorname{LLM}( g) \mid g \Bigl),
\end{align}
where $\mathcal{H}(.)$ is defined as physicians' refinement, 
which ensures symptoms are standard (\textit{e.g.}, mapping ``chest hurt'' to standardized ``chest pain'' or ``angina'').

\noindent\textbf{Disease matching.}
Conditioned on the refined symptom set $\mathcal{\boldsymbol{S}}$ and the clinical guideline $g$, LLM generates a list of differential diagnoses to every single symptom $\mathcal{S}$. The physician then refines this list against guidelines:
\begin{equation}
    \mathcal{\boldsymbol{D}'} = \bigcup_{\mathcal{S}\in\boldsymbol{S}}\mathcal{H}\Bigl( \operatorname{LLM}(g, \mathcal{S})\mid g, \mathcal{S}\Bigr),
\end{equation}
where $\boldsymbol{D'}$ is a set that includes disease name list for each $\mathcal{S}$.
By decoupling disease matching from symptom extraction, we allow physicians to evaluate the validity of each symptom-disease correlation.

\noindent\textbf{Logic completion.}
Based on symptom-diseases pairs (\textit{i.e.}, a core symptom $\mathcal{S}$ paired with its differentials' name list $\mathcal{D'}$) and clinical guideline $g$, LLM extracts clinical evidence $\mathcal{E}$ set and completes differential diagnoses $\mathcal{D}$ set to complete CDRD set of $g$.
For each symptom-disease pair:

\begin{equation}
    \mathcal{E}, \mathcal{D} = \mathcal{H}\Bigl(\operatorname{LLM}( g, (\mathcal{S}, \mathcal{D'}))\mid g, (\mathcal{S}, \mathcal{D'})\Bigl).
\end{equation}
Here, we get the full reasoning data CDRD:
\begin{equation}
    \mathcal{C} = (\mathcal{S}, \mathcal{E}, \mathcal{D}).
\end{equation}
See prompts for the three stages in Appendix \autoref{app:syn-CDRD-prompt}.

While CDRD captures high-quality diagnostic reasoning logic, it suffers from limited scale and lacks dynamic, multi-turn clinical settings. To enable models not only to learn this logic but also to apply it flexibly, we construct Question-Answer data for SFT and multi-turn inquiry data for RL based on CDRD as shown in \autoref{fig:pipe_data} (b, c) separately.

\subsection{QA Pair for SFT}
We use CDRD as a seed to construct QA dataset $\boldsymbol{P}_{\text{SFT}}$ via a two-step process: (i) answer formulation
, based on CDRD, and (ii) question synthesis, based on the answer.

\noindent\textbf{Answer formulation.}
We first parse CDRDs into discrete entries $\boldsymbol{e}$, where $e$ is a partial segment from either $\mathcal{E}$ or $\mathcal{D}$ in a CDRD.
 Each entry $e$ is semantically complete.
To increase expression diversity, we rewrite each entry $e$ into a fluent response $A$ using LLM:
\begin{equation}
    A = \operatorname{LLM}\bigl(e\bigr).
\end{equation}

\noindent\textbf{Question synthesis.}
For each answer $A$, we synthesize a corresponding question $Q$ to form a complete QA pair:
\begin{equation}
    Q =\operatorname{LLM}\bigl( A\bigr).
\end{equation}
These operations are applied to the entire CDRD, yielding the final SFT dataset $\boldsymbol{P}_{\text{SFT}} = \{(Q, A)\}$. Refer to Appendix \autoref{app:syn-QA-prompt} for its synthesis prompts.

\subsection{Multi-turn Inquiry for RL}
\label{inquiry data}
While QA data instill diagnostic reasoning logic, clinical inquiry skills are not covered. 
So we propose a dual-agent comprising a physician agent $\pi_{\text{phy}}$ and a patient agent $\pi_{\text{pat}}$, which takes iterative interactions to synthesize reasoning and inquiries based on patient profiles $\boldsymbol{R}$ and CDRDs $\boldsymbol{C}$.

Specifically, an inquiry dialogue starts with the main complaints specified in $R$, and $\pi_{\text{phy}}$ generates reasoning and inquiry referring to $\mathcal{C}$. We structure the reasoning and inquiry to a template. It consists of six reasoning steps and the final inquiry (See Appendix \autoref{app:inquiry-data-demo} and \autoref{app:syn-inquiry-prompt}, for reasoning-inquiry demos and their synthesis prompts, separately):

\noindent \textbf{Known Information}: The patient's information summary from history dialogue.

\noindent \textbf{User's Intention}: Intentions underlying the query.

\noindent \textbf{Provided Information}: Expressed inquiry summary.

\noindent \textbf{Diagnoses}: Based on the inquiry history, the model generates initial diagnostic hypotheses.

\noindent \textbf{Information to be Collected}: Clinical information required to clarify diagnoses and guide decision-making.

\noindent \textbf{Response Strategy}: Before inquiry, model reasons the most necessary information to acquire.

\noindent \textbf{Inquiry}: Diagnoses or follow-up questions that model presents.

Actually, an inquiry dialogue can be seen as a particular trajectory $\tau$ sampled from the CDRD $\mathcal{C}$, influenced by patient's constraints in $R$:
\begin{equation}
    \tau = (p_1, d_1, p_2, d_2, \dots, p_k, d_k \mid \mathcal{C}, R),
\end{equation}
where $p_k$ represents the $\pi_{\text{pat}}$'s query, and $d_k$ is the $\pi_{\text{phy}}$'s reasoning and inquiry at round $k$.

\section{Dr. Assistant}
Dr. Assistant's training pipeline consists of two stages, as shown in \autoref{fig:pipe_data} (d): Supervised Fine-Tuning (SFT), which equips the model with preliminary clinical diagnostic reasoning logic, and Reinforcement Learning (RL), which further enhances the model's clinical diagnostic inquiry skills.

\paragraph{SFT with $\boldsymbol{P}_{\text{SFT}}$.}
Our model $\pi_{\theta}$ is first fine-tuned via supervised learning on $\boldsymbol{P}_{\text{SFT}}$ to acquire basic diagnostic reasoning logic, optimized with the standard negative log-likelihood loss:
\begin{equation}
    \mathcal{L}_{\text{SFT}}
    = - \mathbb{E}_{(q, a) \sim \boldsymbol{P}_{\text{SFT}}}
        \Biggl[
            \sum_{t=1}^{T}
            \log \pi_{\theta}\bigl(a_t \mid q, a_{<t}\bigr)
        \Biggr].
\end{equation}

\paragraph{RL with multi-turn inquiry.}
Although model learned clinical diagnostic reasoning logic in SFT, its responses are confined to single, static QA pattern, and have limited clinical reasoning and inquiry skills. So we conduct Reinforcement Learning on inquiry dataset $\{\tau\}$, enhancing the model with clinical diagnostic reasoning and inquiry.

Specifically, we design a composite reward function. Given a ground-truth reasoning and inquiry $d_t$ at round $t$ of an inquiry dialogue $\tau$ and the trained model's response $\hat{d}_t$, the step-wise reward is defined as:
\begin{equation}
    R_{\text{step}}(\hat{d}_t, d_t, \mathcal{C})
    = R_{\text{comp}}(\hat{d}_t, d_t)
      - R_{\text{div}}(\hat{d}_t, \mathcal{C}).
\end{equation}
Here, $R_{\text{comp}}$ measures: \textit{clinical reasoning} $R_{\text{comp-r}}$, \textit{and inquiry skills} $R_{\text{comp-i}}$, while $R_{\text{div}}$ constraints \textit{fidelity to CDRD}.

\noindent\textbf{a. Clinical reasoning and inquiry skills.}  
$R_{\text{comp}}$ evaluates the model's performance by scoring the similarity $r_i$ between its output and the ground‑truth across seven parts related to diagnostic reasoning and inquiry: 
1) \textit{Reasoning list} $\boldsymbol{r}_{\text{reason}}$ including Known Information $r_0$, User Intention $r_1$, Provided Information $r_2$,  Diagnoses $r_3$, Information to be Collected $r_4$ and Response Strategy $r_5$,  
2) \textit{Inquiry list} $\boldsymbol{r}_{\text{inquiry}}$ including the final inquiry $r_6$. Definitions of the seven parts are in {\autoref{inquiry data}}.

$r_{i}$ is scored by another LLM-judge (see judge prompt in Appendix \autoref{app:RL-laaj-prompt-comp}), ranging from 0 to 10. A higher alignment with the ground-truth yields a higher score. $R_{\text{comp}}$ is derived as:
\begin{equation}
\label{eq:rm1}
    \begin{split}
    R_{\text{comp}}(d_t, \tau) & = R_{\text{comp-r}} + R_{\text{comp-i}} \\
    & = w_{\text{reason,a}} \boldsymbol{r}_{\text{reason,a}} + w_{\text{reason,b}} \boldsymbol{r}_{\text{reason,b}} \\
    &+ w_{\text{inquiry}} \boldsymbol{r}_{\text{inquiry}},
    \end{split}
\end{equation}
where $\boldsymbol{r}_{\text{reason,a}} = [r_0, r_1, r_2]^\top$, $\boldsymbol{r}_{\text{reason,b}} = [r_3, r_4, r_5]^\top$, and $\boldsymbol{r}_{\text{inquiry}}$ is the score vector for inquiry items. 
$w_{\text{reason,a}}$, $w_{\text{reason,b}}$ and $w_{\text{inquiry}}$ are weight coefficients, set according to the relative importance of each dimension in clinical decision-making. See \autoref{app:parametersetting} for parameter settings.

\noindent\textbf{b. Fidelity to CDRD.} $R_{\text{div}}$ penalizes deviations from the source CDRD (\textit{i.e.}, the specific CDRD used to generate the ground-truth trajectory $\tau$ that serves as the target in this optimization step): 
\begin{equation}
\label{eq:rm2}
      R_{\text{div}}(d_t, \mathcal{C}) = \lambda n,
\end{equation}
where $\lambda$ is the weight coefficient and $n$ is the number of already collected information (or yet to be collected) and diagnoses that are \textbf{not} in CDRD, determined by LLM-judge (see judge prompt in Appendix \autoref{app:RL-laaj-prompt-div}).

We employ DAPO \cite{yu2025dapoopensourcellmreinforcement}, which samples
output group $\{o_{i}\}_{i=1}^{G}$ for each prompt and optimizes the model $\pi_{\theta}$ by the following objective:
\begin{equation}
\small
    \begin{split}
    \mathcal{J}_{\text{DAPO}}(\theta) = \mathbb{E} \Bigg[ &\frac{1}{\sum_{i=1}^G |o_i|} \sum_{i=1}^{G} \sum_{t=1}^{|o_i|} \min \Big(r_{i,t}(\theta) \hat{A}_{i,t}, \\
    &\operatorname{clip}(r_{i,t}(\theta), 1-\epsilon_{low}, 1+\epsilon_{high}) \hat{A}_{i,t} \Big) \Bigg],
    \end{split}
\end{equation}
where $r_{i,t}(\theta) = \frac{\pi_{\theta}(o_{i,t} \mid q,o_{i},<t)}{\pi_{\text{old}}(o_{i,t} \mid q,o_{i},<t))}$ is the probability ratio and $o_i$ is the length of trajectory $\tau_i$.
$\hat{A}_{i,t} = \frac{R_{i}-mean(\{R_{i}\}_{i=1}^G)}{std(\{R_{i}\}_{i=1}^{G})}$, is advantage of the $i$-th response is calculated by normalizing the group-level rewards $\{R_{i}\}_{i=1}^G$.

\section{Experiments}

\subsection{Experimental Settings}
\noindent\textbf{Model and baselines.} 
Our Dr. Assistant is trained on Qwen3-14B \cite{qwen3technicalreport}. We compare it against: 1) \textit{Open-source Models}: including general-purpose LLMs: Qwen3-32B \cite{qwen3technicalreport} , DeepSeek-R1-Distill-14B \cite{deepseekai2025deepseekr1incentivizingreasoningcapability}, Baichuan-M2-32B \cite{2}, Llama3-OpenBioLLM-70B \cite{OpenBioLLMs} and medical-specialized LLMs (HuatuoGPT-o1-72B \cite{huatuogpt-o1} , Doctor-R1-8B \cite{lai2025doctorr1masteringclinicalinquiry}, Med42-v2-8B \cite{med42v2}. 2) \textit{Closed-source Models}: This includes leading proprietary models: GPT-5 \cite{gpt-5} , Gemini 2.5 Pro \cite{gemini25-pro}, Gemini-2.5-Flash \cite{gemini25-flash} and Grok-4-Fast \cite{grok-4-fast}. Details of parameter settings are in \autoref{app:parametersetting}.

\noindent\textbf{Training datasets.} 
We construct 11 CDRDs covering $\sim$60 diseases based on internal guidelines for training.
For each CDRD, 1000 QA pairs are synthesized, totaling 11000 data for SFT.
Additionally, we construct 36688 turns of multi-turn inquiry dialogues for RL.
All training data is in Chinese.

\noindent\textbf{Evaluation.} 
Existing benchmarks, such as HealthBench \cite{arora2025healthbench} and MAQuE \cite{22}, do not concentrate on dynamic clinical diagnostic inquiry reasoning. HealthBench contains general medical consultations, most of which are not multi-turn dialogues. MAQuE does not consider the mutual influence between the diagnostic inquiry process and the final diagnoses.
However, inspired by these evaluations \cite{dou2024integrating, arora2025healthbench, 22}, we construct a benchmark for clinical diagnostic inquiry. It considered the dynamic interaction and influence between the inquiry process and final diagnoses, which encompasses 242 patient profiles (cases) across nine sub-departments, including neurology, and 147 rounds of clinical inquiry dialogue records, as shown in \autoref{fig:bench_distribution}. See demos in Appendix \autoref{app:patient-profile-demo} and \autoref{app:real-inquiry-history-demo}.

We evaluate the model via both \textbf{ICD-10 \cite{icd-10} matching} for diagnostic reasoning and \textbf{physician satisfaction for inquiry. 
}

\begin{figure}[t!]
    \centering
    \begin{subfigure}{0.25\textwidth}
        \includegraphics[width=\linewidth]{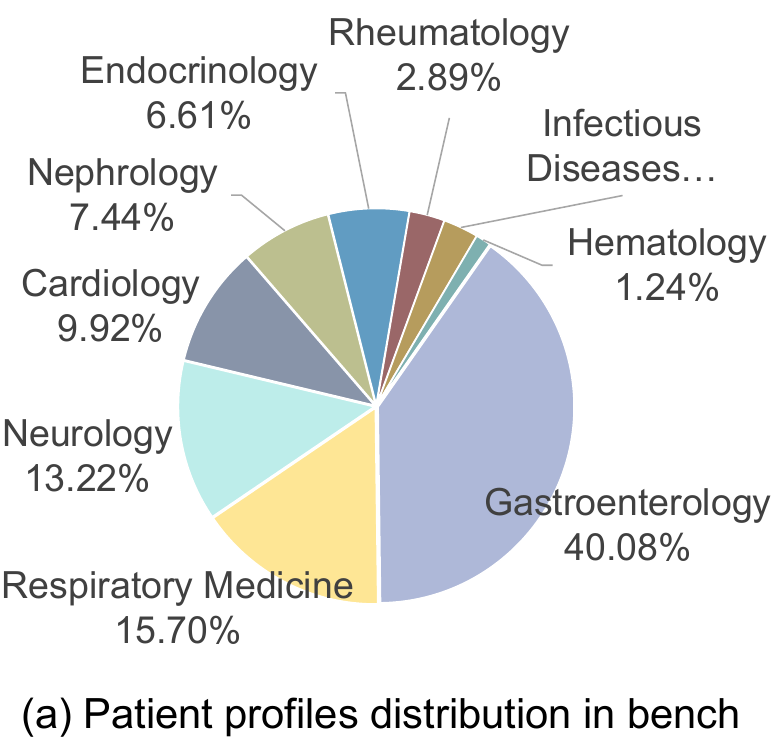}
    \end{subfigure}%
    \begin{subfigure}{0.25\textwidth}
        \includegraphics[width=\linewidth]{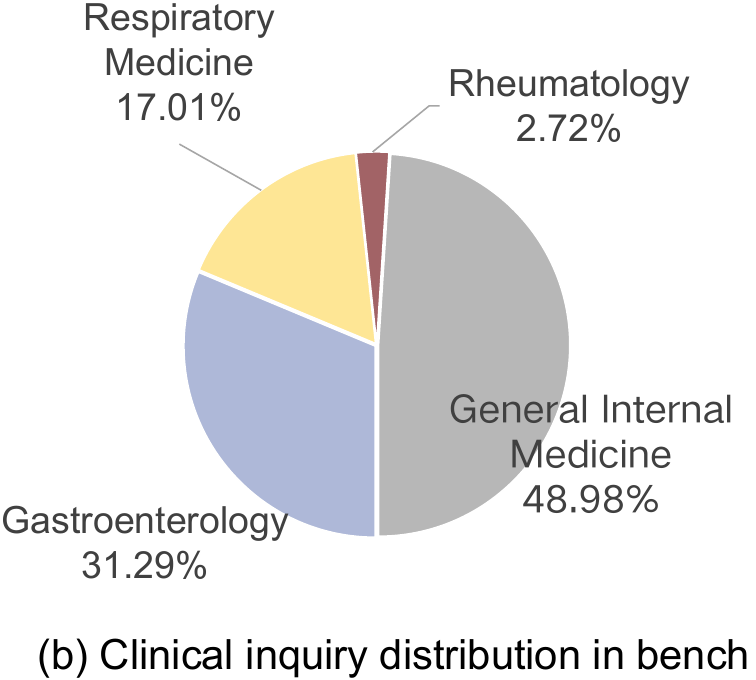}
    \end{subfigure}
    \caption{Data distribution of our benchmark. One includes 242 patient profiles across 7 sub-departments (a). The other includes 147 rounds of clinical inquiry dialogues (b). ``General Internal Medicine'' means the inquiries cover multiple sub-departments.}
    \label{fig:bench_distribution}
\end{figure}

\subsection{ICD Matching for Diagnostic Reasoning}
Since diagnostic hypotheses in reasoning directly determine inquiry direction, as shown in \autoref{fig:teaser} (a), we design an evaluation where one model, acting as a patient based on a given patient profile, interacts with the model-under-test (physician model). In each turn, the physician model outputs diagnostic hypotheses (the core reasoning) and inquiry. 
We compare diagnostic hypotheses at round 3, 5 and 10, with GT in the patient profile, to assess the model's diagnostic reasoning ability and inquiry direction. Results for rounds 5 are shown in \autoref{tab:oas_icd_results}. Results for rounds 3, 10 are also provided in Appendix \autoref{tab:prf_round3510}. Patient profiles used in bench are different from those in \autoref{inquiry data}. For more details, see Appendix \autoref{app:icd-10-metric}.

\noindent\textbf{ICD-10 matching.} 
We map the diagnostic hypotheses of both model and GT to ICD-10 (International Classification of Diseases) \cite{icd-10} codes, $\mathcal{P}$ and $\mathcal{G}$ respectively by LLM.
We then calculate their similarity as formalized in Appendix \autoref{alg:icd_matching}. 
In clinical practice, the \textbf{ICD-Recall} of diagnostic hypotheses is positively correlated with the \textit{reasoning} and its \textit{whole inquiry strategy}, which is our \textbf{main target}. But we still include precision for reference.

\noindent\textbf{ICD-Recall ($R_{\text{ICD}}$).} This is our \textbf{primary metric}, measuring the coverage of ground truth diagnoses.
\begin{equation}
    R_{\text{ICD}} = \frac{1}{|\mathcal{G}|} \sum_{g \in \mathcal{G}} \max_{p \in \mathcal{P}} \operatorname{Sim}_{\text{ICD}}(p, g).
\end{equation}

\noindent\textbf{ICD-Precision ($P_{\text{ICD}}$).} It reflects diseases requiring exclusion via further inquiry in diagnostic hypotheses, while not in GT:
\begin{equation}
    P_{\text{ICD}} = \frac{1}{|\mathcal{P}|} \sum_{p \in \mathcal{P}} \max_{g \in \mathcal{G}} \operatorname{Sim}_{\text{ICD}}(p, g).
\end{equation}

As shown in \autoref{tab:oas_icd_results}:
(1) Overall, Dr. Assistant achieves an advance in ICD-Recall, improving by 6.70\% relative to the best closed-source model, Gemini-3-Pro-Preview (0.5066 v.s. 0.4748). This demonstrates its effectiveness.
(2) Compared to open-source general-purpose models, Dr. Assistant outperforms Qwen3-32B by 16.30 \% relatively (0.5066 v.s. 0.4356). 
This is because Dr. Assistant learned diagnostic reasoning logic and inquiry skills to infer patient's disease from a core symptom.
(3) Compared to open-source medical models, Dr. Assistant achieves state-of-the-art in ICD-Recall, even surpassing HuatuoGPT-o1-72B by 13.59\% relatively (0.5066 v.s. 0.4460). It is because previous medical models focus on domain knowledge but lack clinical diagnostic reasoning and inquiry skills. In RL stage, Dr. Assistant internalized reasoning and inquiry skills, thereby driving the inquiry dialogue to an optimal trajectory.
(4) Dr. Assistant is not the best in ICD Precision (candidate diagnosis count). As shown in \autoref{tab:oas_icd_results}, its precision is 0.4717, compared to 0.5211 for HuatuoGPT-o1-72B and 0.4842 for GPT-5. Dr. Assistant averages 2.1983 candidates per case, slightly higher than GPT 5's 1.7851. This confirms Dr. Assistant's high recall without candidate flooding. But if the correct candidate is missing (low ICD-Recall), even a small candidate list may cause missed diagnosis.

\begin{table}[t]
\centering
\small 
\renewcommand{\arraystretch}{1.2}

\setlength\tabcolsep{3pt}

\renewcommand{\tabularxcolumn}[1]{m{#1} }

\newcolumntype{L}{>{\raggedright\arraybackslash\hsize=1.8\hsize}X} 
\newcolumntype{Y}{>{\centering\arraybackslash\hsize=0.6\hsize}X}
\newcolumntype{G}{>{\color{gray}\centering\arraybackslash\hsize=0.6\hsize}X}

\begin{tabularx}{\columnwidth}{L Y G}
\toprule
\textbf{Model} & 
\textbf{Recall}$^{\dagger}$ $\uparrow$ &
\textcolor{gray}{\textbf{Prec.}} $\uparrow$ \\ 
\midrule

\rowcolor[gray]{0.95}
\multicolumn{3}{l}{\textbf{Open-source (Medical)}} \\
HuatuoGPT-o1-72B                & 0.4460 & \underline{0.5211} \\
Baichuan-M2-32B                 & 0.3687 & 0.4517 \\
Doctor-R1-8B                    & 0.3873 & 0.4800 \\
Llama3-OpenBioLLM-70B                   & 0.2216 & 0.2755\\
Med42-v2-8B                   & 0.2787 & 0.3081 \\
\midrule
\rowcolor[gray]{0.95}
\multicolumn{3}{l}{\textbf{Open-source (Non-medical)}} \\
Qwen3-14B                        & 0.4298 & 0.4600 \\
Qwen3-32B                        & 0.4356 & 0.4660 \\
DeepSeek-R1-Distill-14B          & 0.3913 & 0.4267 \\

\midrule

\rowcolor[gray]{0.95}
\multicolumn{3}{l}{\textbf{Closed-source}} \\
GPT-5                            & 0.4651 & 0.4842 \\
Gemini-2.5-Flash           & 0.4288 & 0.4497 \\
Gemini-3-Pro-Preview                 & \underline{0.4748} & \textbf{0.5333} \\
Grok-4-Fast                 & 0.4284& 0.5159 \\
\midrule
\rowcolor[RGB]{236,244,252}
\textbf{Dr. Assistant} & \textbf{0.5066 } & 0.4717 \\
\bottomrule

\end{tabularx}
\caption{
Average ICD-Recall and Precision, sorted by \textbf{ICD-Recall$^{\dagger}$}. ICD-Recall (denoted by $\dagger$) is the primary criterion for assessing a model's diagnostic reasoning correctness. Results reported here are from round 5. Stability analysis across rounds 3, 5, and 10 is shown in \autoref{tab:icd_sensitivity}. Results for rounds 3, 10 are provided in Appendix \autoref{tab:prf_round3510}. Model performance across secondary departments is detailed in Appendix \autoref{app:tab_secondary-icd}.
}

\label{tab:oas_icd_results}

\end{table}

\begin{table}[t]
\centering
\small 
\renewcommand{\arraystretch}{1.2}
\setlength\tabcolsep{3pt}
\renewcommand{\tabularxcolumn}[1]{m{#1}}
\renewcommand{\arraystretch}{1.2}
\setlength{\tabcolsep}{3pt}
\newcolumntype{T}{>{\raggedright\arraybackslash}X}

\begin{tabularx}{\columnwidth}{T ccc}
\toprule
\textbf{Model} & 
\textbf{Mean} & 
\textbf{Std} & 
\textbf{CV (\%)} \\ 
\midrule

\rowcolor[gray]{0.95}
\multicolumn{4}{l}{\textbf{Open-source (Medical)}} \\
HuatuoGPT-o1-72B          & 0.4431 & 0.0058 & 1.30 \\
Baichuan-M2-32B           & 0.3740 & 0.0100 & 2.67 \\
Doctor-R1-8B              & 0.3817 & 0.0064 & 1.66 \\
Llama3-OpenBioLLM-70B     & 0.2031 & 0.0162 & 8.00 \\
Med42-v2-8B               & 0.2870 & 0.0318 & 11.07 \\
\midrule

\rowcolor[gray]{0.95}
\multicolumn{4}{l}{\textbf{Open-source (Non-medical)}} \\
Qwen3-14B                 & 0.4242 & \underline{0.0053} & \underline{1.24} \\
Qwen3-32B                 & 0.4315 & 0.0138 & 3.19 \\
DeepSeek-R1-Distill-14B   & 0.3963 & 0.0172 & 4.33 \\
\midrule

\rowcolor[gray]{0.95}
\multicolumn{4}{l}{\textbf{Closed-source}} \\
GPT-5                     & 0.4555 & 0.0085 & 1.87 \\
Gemini-2.5-Flash          & 0.4375 & 0.0120 & 2.74 \\
Gemini-3-Pro-Preview      & \underline{0.4690} & 0.0100 & 2.14 \\
Grok-4-Fast               & 0.4302 & 0.0099 & 2.31 \\
\midrule

\rowcolor[RGB]{236,244,252}
\textbf{Dr. Assistant}     & \textbf{0.5060} & \textbf{0.0007} & \textbf{0.14} \\
\bottomrule
\end{tabularx}
\caption{Sensitivity analysis of ICD-Recall across different inquiry turns (3, 5, 10). Mean, standard deviation, and Coefficient of Variation (CV\%) are reported. Dr. Assistant shows consistent performance with minimal variance, confirming that its results are not dependent on a specific turn threshold.}
\label{tab:icd_sensitivity}
\end{table}

\subsection{Physician Satisfaction}

\begin{figure}[t] 
    \centering
    \includegraphics[width=0.5\textwidth]{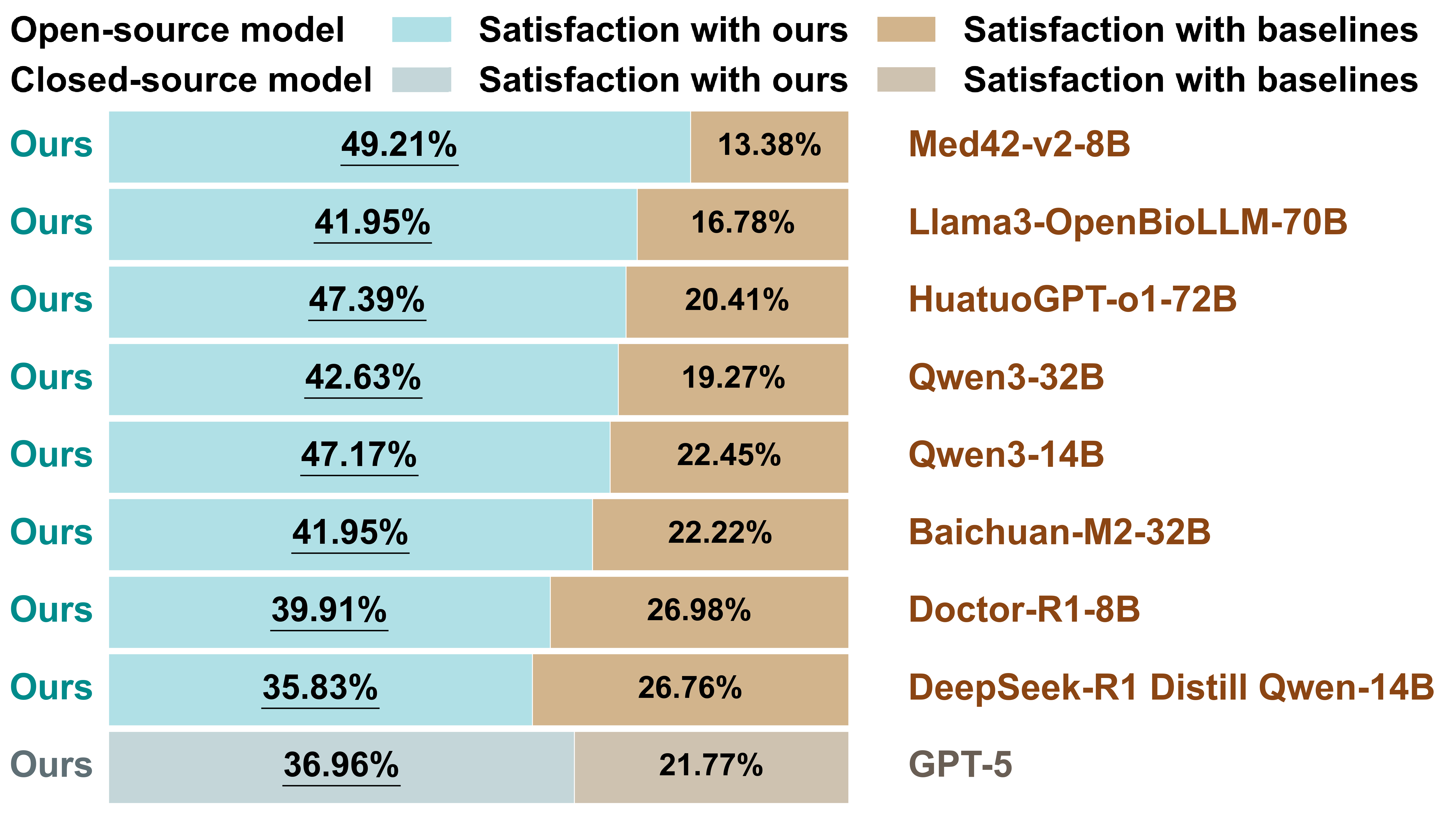}  
    \caption{Physician's satisfaction. For a clearer display of comparison, tie cases are excluded from this figure.}
    \label{fig:satisf-baseline}
\end{figure}

To evaluate model inquiries in real application settings, we had the model produce inquiries based on real clinical inquiry records (147 rounds), submitted them to physicians for evaluation. We invited 11 physicians with 5 to 8 years of clinical experience (average around 6 years) and specialties including surgery, gynecology, internal medicine, and traditional Chinese medicine to participate in the evaluation.
The evaluation criteria consist of two parts: \textit{relevance} and\textit{ expertise}, where relevance serves as the prerequisite for winner. Physicians must determine a winner of inquiry-couple (ours, baseline), or declare a tie. See Appendix \autoref{app:satisfaction} for details.
Each evaluation was done by at least two physicians in a double blind manner, with a Cohen's Kappa of 0.955. A third senior physician resolved any disagreement until consensus. All physicians saw the full results only after the evaluation.

As shown in \autoref{fig:satisf-baseline}:
(1) Dr. Assistant gets higher physician satisfaction compared to open-source models. It shows improvements of 35.83\% and 26.78\% over the medical model Med42-v2-8B and HuatuoGPT-o1-72B, respectively. 
Dr. Assistant maintains focus on the diagnostic target. It stems from its clear diagnostic reasoning logic, which enables it to deliver more valuable inquiries throughout the clinical process.
See case study in \autoref{case-study}.
(2) Dr. Assistant advances GPT-5 by 15.19\% (36.96 v.s. 21.77). This highlights that complex clinical inquiry requires precise diagnostic reasoning. Dr. Assistant integrates diagnostic reasoning logic during RL, achieving superior physician satisfaction.

\begin{table}[h]
\centering
\small

\renewcommand{\arraystretch}{1.2} 
\setlength\tabcolsep{3pt}

\renewcommand{\tabularxcolumn}[1]{m{#1}}

\newcolumntype{L}{>{\raggedright\arraybackslash\hsize=1.3\hsize}X} 
\newcolumntype{C}{>{\centering\arraybackslash\hsize=0.6\hsize}X}
\newcolumntype{D}{>{\centering\arraybackslash\hsize=0.5\hsize}X}

\begin{tabularx}{\columnwidth}{L C C D}
\toprule
\textbf{Benchmark} & 
\textbf{Qwen3-14B} & 
\textbf{Dr. Assistant} & 
\textbf{Retention} \\ 
\midrule

\rowcolor[gray]{0.95}
\multicolumn{4}{l}{\textit{\textbf{CMMLU (Medical, Chinese)}}} \\
Anatomy                & 0.8649 & 0.8784 & \textbf{101.56\%} \\
Professional Medicine  & 0.7952 & 0.8032 & \textbf{101.01\%} \\
Nutrition              & 0.8207 & 0.8276 & \textbf{100.84\%} \\
Clinical Knowledge     & 0.7890 & 0.7932 & \textbf{100.53\%} \\
TCM                    & 0.8324 & 0.8324 & \textbf{100.00\%} \\
College Medicine       & 0.8828 & 0.8791 & {99.58\%}  \\
Genetics               & 0.7727 & 0.7670 & {99.26\%}  \\
Virology               & 0.8639 & 0.8580 & {99.32\%}  \\

\midrule

\rowcolor[gray]{0.95}
\multicolumn{4}{l}{\textit{\textbf{MMLU-Pro (English)}}} \\
Zero-shot       & 0.6710 & 0.6799 & \textbf{101.33\%} \\
Five-shot       & 0.6724 & 0.6686 & 99.43\% \\

\midrule

\rowcolor[gray]{0.95}
\multicolumn{4}{l}{\textit{\textbf{MedQA (English)}}} \\
Zero-shot       & 0.7021 & 0.6897 & 98.23\% \\
Five-shot       & 0.6960 & 0.6952 & 99.89\% \\

\midrule
\rowcolor[RGB]{236,244,252}
\textbf{Average (Overall)} & {0.7803} & {0.7810} & \textbf{100.10\%} \\
\bottomrule
\end{tabularx}
\caption{
Performance on general medical benchmarks. Knowledge retention rate measures the model's ability to preserve original medical knowledge after post training. Post-training on vertical tasks often leads to severe catastrophic forgetting. Our model retains about 100 percent of its general medical knowledge, confirming that our training method safely preserves the model's original medical knowledge.}

\label{tab:score_on_qa}
\end{table}

\subsection{General Medical Knowledge Retention}

To assess the general knowledge retention of Dr. Assistant, we conduct a comparative analysis of its performance against Qwen3-14B on the general knowledge benchmarks CMMLU (0/5-shot) \cite{li2024cmmlu},  MMLU-Pro (0/5-shot) \cite{wang2024mmlupro}, and MedQA \cite{jin2021medqa}. For CMMLU, we select eight medical-related subsets for evaluation. Results are in \autoref{tab:score_on_qa}.

Overall, Dr. Assistant maintains medical knowledge to Qwen3-14B (0.7810 v.s. 0.7803), demonstrating no significant knowledge forgetting. Notably, on the eight Chinese medical subsets of CMMLU \cite{li2024cmmlu}, Dr. Assistant achieves equal or superior performance to Qwen3-14B in five of them, further validating the effectiveness of our training method.

\begin{table}[h]
\centering
\small 

\label{tab:ablation_results}

\renewcommand{\arraystretch}{1.2} 
\setlength\tabcolsep{3pt}

\renewcommand{\tabularxcolumn}[1]{m{#1}}

\newcolumntype{L}{>{\raggedright\arraybackslash\hsize=1.3\hsize}X} 
\newcolumntype{Y}{>{\centering\arraybackslash\hsize=0.9\hsize}X}
\newcolumntype{G}{>{\color{gray}\centering\arraybackslash\hsize=0.9\hsize}X}

\begin{tabularx}{\columnwidth}{L Y Y G}
\toprule
\textbf{Method} & 
\textbf{Satisfaction} &  
\textbf{Recall}$^{\dagger}$ $\uparrow$ & 
\textcolor{gray}{\textbf{Prec.}} $\uparrow$ \\ 
\midrule

\rowcolor[RGB]{236,244,252}
\textbf{Dr. Assistant} & \textbf{-0.00\%} & \textbf{0.5066} & 0.4717 \\
\hline
w/o SFT                 & -18.82\% & \underline{0.5024} & 0.4475 \\
w/o RL                  & -6.80\%  & 0.3927 & \underline{0.5144} \\
w/o $R_{\text{div}}$    & -8.62\%  & 0.4856 & \textbf{0.5192} \\
w/o $R_{\text{comp-r}}$ & \underline{-5.50\%}  & 0.4893 & 0.4830 \\

\bottomrule
\end{tabularx}
\caption{
Ablation study on \textbf{Dr. Assistant}. Satisfaction is calculated as the difference in winning percentage compared to Dr. Assistant.
}
\label{ablation_study}
\end{table}

\subsection{Ablation Study}

We evaluate the contributions of SFT, RL, and the reward function to Dr. Assistant on our benchmark, with results summarized in \autoref{ablation_study}. 

The removal of SFT and RL leads to performance drops of 0.84\% and 29.00\% relative to the full on ICD-Recall separately, and drops of 18.82\% and 6.80\% on satisfaction separately. SFT improves the model's clinical diagnostic logic, while RL further reinforces reasoning and inquiry skills. Notably, performance degrades less when SFT is omitted than when RL is. This suggests SFT tends to solidify the model's response patterns around the tuning data, which may constrain its adaptability to prior tasks. And RL not only guides the model beyond these fixed patterns but also further refines and expands its learned capabilities. \cite{DBLP:journals/corr/abs-2507-05386}.

Furthermore, we individually removed the $R_{\text{div}}$ and $R_{\text{comp-r}}$ reward. This leads to performance declines of 4.32\% and 3.54\% on ICD-Recall, respectively, and 8.62\% and 5.50\% on satisfaction, respectively, demonstrating their effectiveness. The $R_{\text{div}}$ component serves to penalize the generation of incorrect diagnostic information, improving output quality. The $R_{\text{comp-r}}$ function rewards the model's clinical reasoning, which is similar to GT.

\subsection{Easy and Hard Diagnostic Inquiry}

\begin{table}[h]
\centering
\small
\setlength{\tabcolsep}{3pt}
\renewcommand{\arraystretch}{1.2}

\renewcommand{\tabularxcolumn}[1]{m{#1}}
\newcolumntype{L}{>{\raggedright\arraybackslash\hsize=1.6\hsize}X} 
\newcolumntype{C}{>{\centering\arraybackslash\hsize=0.7\hsize}X}

\begin{tabularx}{\linewidth}{L C C}
\toprule
\textbf{Model} & \textbf{Diag.=1} & \textbf{Diag.>1} \\
\midrule

\rowcolor[gray]{0.95}
\multicolumn{3}{l}{\textit{Open-source (medical)}} \\
HuatuoGPT-o1-72B        & 0.5248 & 0.4461 \\
Baichuan-M2-32B         & 0.4139 & 0.3688 \\
Doctor-R1-8B            & 0.4733 & 0.3890 \\
Med42-v2-8B             & 0.2938 & 0.2788 \\
Llama3-OpenBioLLM-70B   & 0.2240 & 0.1867 \\
\cmidrule(lr){1-3}

\rowcolor[gray]{0.95}
\multicolumn{3}{l}{\textit{Open-source (general)}} \\
Qwen3-14B               & 0.5030 & 0.4316 \\
Qwen3-32B               & 0.5307 & 0.4356 \\
DeepSeek-R1-Distill-14B & 0.4198 & 0.3913 \\
\midrule

\rowcolor[gray]{0.95}
\multicolumn{3}{l}{\textit{Closed-source}} \\
GPT-5                   & 0.5287 & 0.4652 \\
Gemini 2.5 Flash        & 0.5192 & 0.4289 \\
Grok-4-Fast             & 0.4821 & 0.4092 \\
Gemini-3-Pro-Preview    & \textbf{0.5455} & \underline{0.4749} \\
\midrule

\rowcolor[RGB]{236,244,252}
\textbf{Dr. Assistant} & {\underline{0.5426}} & \textbf{0.5067} \\

\bottomrule
\end{tabularx}

\caption{Model performance (ICD-Recall) stratified by diagnostic complexity: single-disease cases (Diag.=1, easy) vs. multi-disease cases (Diag.>1, hard). We provide detailed Diag.=1$\sim$6 in Appendix \autoref{app:tab_count_comparison}.}
\label{tab:count_1_vs_m_wide}
\vspace{-2mm}
\end{table}

 We stratified patient profiles by disease count (1 to 6) into single disease (easy) and multi disease (hard) cases. Multi disease cases form a multi-label diagnostic task. The model collects information via multi-turn inquiry and outputs diagnostic hypotheses, each matched against ground truth. All cases in our test set can be diagnosed through inquiry. Cases requiring additional biopsy or exams are excluded.
 As shown in \autoref{tab:count_1_vs_m_wide}, compared to Baichuan-M2-32B and GPT-5, Dr. Assistant achieves disease-recall improvements of 31.09\% and 2.80\% (0.5426 v.s. 0.4139 and 0.5426 v.s. 0.5278) on single-disease tasks, and 37.39\% and 8.92\% (0.5067 v.s. 0.3688 and 0.5067 v.s. 0.4652) on multi-disease tasks, respectively. Dr. Assistant not only achieves competitive performance on single-disease tasks but also demonstrates superior diagnostic performance in hard (multi-disease) settings. It is because Dr. Assistant stays on the diagnostic target. It stems from its clear diagnostic reasoning logic, which enables it to deliver more valuable inquiries throughout the clinical process.
We also analyze the model's performance across different departments, as shown in \autoref{department-performance}.

\section{Conclusion}
To address the limitations of traditional CDSSs and enhance the diagnostic reasoning capability of LLMs in clinical diagnostic inquiry, we propose (1) a structured Clinical Diagnostic Reasoning Data (CDRD) format along with a construction pipeline to capture abstract clinical reasoning logic, and (2) the Dr. Assistant model, which integrates clinical reasoning and inquiry skills through a two-stage training strategy of Supervised Fine-Tuning followed by Reinforcement Learning with a tailored reward function. Evaluated on a dedicated diagnostic reasoning and inquiry benchmark, Dr. Assistant outperforms open-source models and achieves competitive performance compared to closed-source models, offering a practical solution for clinical diagnostic inquiry guidance.

\section{Limitation} 
Our study has several limitations. First, the construction of Clinical Diagnostic Reasoning Data (CDRD) relies on physician refinement, which introduces manual effort and may limit scalability. Although we used a three stage pipeline to ensure quality, this process remains resource intensive. Second, our evaluation benchmark, while built from real clinical cases and dialogues, is small in scale and covers a limited number of secondary departments. Third, the reward function used in reinforcement learning, though designed with clinical dimensions in mind, may not fully capture all nuances of real diagnostic reasoning. Future work could explore more automated, data driven reward modeling or incorporate direct clinical outcome feedback. Fourth, as shown in \autoref{tab:oas_icd_results}, the precision of Dr. Assistant is not the highest. Future work could further optimize candidate prioritization. Fifth, our test set only contains 242 cases and 147 rounds of clinical inquiry records, which is a small scale. Future work should validate on larger datasets and systematically analyze model performance across different numbers of inquiry turns. Finally, our experiments focus primarily on diagnostic inquiry within a controlled dialogue framework. The model's performance in fully open ended clinical conversations or integration with live electronic health record systems remains to be tested.

\section{Acknowledgments}
We are grateful to Junfeng Wang and Bo Yuan for their leadership and support on this project. We extend our special thanks to Jianwei Lv and other colleagues for their helpful comments on earlier drafts of the paper. We also sincerely appreciate the strong support from our professional medical team in both data acquisition and model evaluation. All listed individuals are with Baidu Inc.


\bibliography{custom}

\clearpage

\appendix

\newpage

\section{Parameter Setting}
\label{app:parametersetting}

\noindent\textbf{Data construction.} 
We use DeepSeek-V3.2 \cite{deepseekai2025deepseekv32} for constructing CDRD data. For constructing QA pairs and multi-turn inquiry dialogue data, we employ Qwen3-235B-A22B \cite{qwen3technicalreport}. The model selection is determined through manual validation on a small set of test data.

\noindent\textbf{SFT.}
The training epoch is set to 4, and a cosine annealing strategy is adopted, with a maximum learning rate of 5e-6. The warm-up ratio is set to 0.0,3 and the regularization coefficient is set to 0.01. SFT training process is conducted on 32 A800 GPUs. 

\noindent\textbf{RL.}
Our RL training is based on the VeRL framework \cite{sheng2024hybridflow} with Qwen3-32B \cite{qwen3technicalreport} as the LLM-judge in our reward function. The learning rate is set to ($1 \times 10^{-6}$), clipping parameters are ($\epsilon_{low} = 0.2$) and ($\epsilon_{high} = 0.28$), the rollout number is 4, and the training batch size is 512. The entire reinforcement learning process was run for 185 steps.
$R_{\text{comp}}$'s (\cref{eq:rm1}) weights are set to $w_\text{reason,a} = 0.1$, $w_\text{reason,a} = 0.3$ and $w_\text{inquiry} = 0.6$. $R_{\text{div}}$'s (\cref{eq:rm2}) weight is set to $\lambda = 5$.
The RL training process is conducted on 8 H800 GPUs, with an additional 16 H800 GPUs used for the LLM-judge.

\noindent\textbf{Evaluation.} To evaluate the ICD-Recall, ICD-Precision and Physician Satisfaction metrics, we set greedy decoding (with temperature = 0) for all models to ensure deterministic and comparable generation outputs. The prompts in \autoref{app:eval-prompt-icd-gen} and \autoref{app:eval-prompt-satisfaction-gen} are used to generate responses for metric computation.

\section{Evaluation}
\label{app:evaluation}

Dr. Assistant is for clinical diagnostic reasoning and inquiry guidance (internal-medicine in Chinese in experiment). However, existing benchmarks are not fully aligned with it. HealthBench \cite{arora2025healthbench} covers general healthcare consultation instead of multi-turn diagnostic inquiry. MAQuE \cite{22}, while involving multi-turn diagnostic inquiry, separates inquiry and diagnostic reasoning. Therefore, it does not align the real-world interdependence between them. 

\subsection{Reasoning-inquiry Loop}
\label{app:loop}

In practice, the reasoning leads to a different trajectory of inquiry. 
Physician issues inquiry based on their diagnostic hypotheses (the reasoning core). Upon receiving an inquiry, the patient responds with new queries. The physician then updates the diagnostic hypothesis based on this new query and continues the dialogue. We show this reasoning-inquiry loop in \autoref{fig:teaser} (a), which ultimately results in different diagnostic conclusions for the same patient. 

Although the reasoning before an inquiry consists of multiple steps as mentioned in \autoref{inquiry data}, their ultimate goal is to verify the diagnostic hypothesis. Including the correct diagnoses within this hypothesis is a prerequisite for proper inquiry.

\subsection{ICD-10 Matching of Clinical Diagnostic Reasoning}
\label{app:icd-10-metric}
Physicians form initial diagnostic hypothesis based on available information and iteratively refine it through targeted inquiry as new information is gathered, ultimately arriving at diagnostic results. The evolving diagnostic hypothesis drives the reasoning of inquiry. Therefore, we compare the model's diagnostic hypothesis with the diagnosis GT.
To simulate dynamic clinical inquiry, we use an LLM as a patient simulator, conditioned on real patient profiles (includes basic medical history, communication style and diagnosis, \textbf{anonymized}). Each inquiry starts from patient simulator's main complaint. At each turn, the evaluated model outputs two parts: a \emph{diagnostic hypothesis} and an \emph{inquiry} to the patient. After five rounds, we compare the model's last diagnostic hypotheses with the diagnoses in the patient profile. Average ICD-Recall and ICD-Precision of each baseline are reported in \autoref{tab:oas_icd_results}.

\noindent\textbf{ICD-10 matching.} 
ICD-10 (International Classification of Disease) \cite{icd-10} codes' leading letter indicates major chapter and subsequent digits specify granular categories. We map the diagnosis results by LLM from both the model and the ground truth to ICD-10 codes, $\mathcal{P}$ and $\mathcal{G}$, respectively. 
We then calculate the similarity between them using $\operatorname{Sim}_{\text{ICD}}(p, g)$, as formalized in \textbf{\autoref{alg:icd_matching}}. 
Based on this similarity, we compute two metrics:

\noindent\textbf{ICD-Recall ($R_{\text{ICD}}$).} This is our \textbf{primary metric}, measuring the coverage of ground truth diagnoses.
\begin{equation}
    R_{\text{ICD}} = \frac{1}{|\mathcal{G}|} \sum_{g \in \mathcal{G}} \max_{p \in \mathcal{P}} \operatorname{Sim}_{\text{ICD}}(p, g)
\end{equation}
\textbf{Reason for ICD-Recall priority:} 
In clinical diagnostic reasoning, missing right diagnoses means the model loses critical diagnostic clues, leading to skewed inquiry and results. More importantly, it poses a severe safety risk as the true condition remains unaddressed. Conversely, a false alarm can be ruled out by a physician.

\noindent\textbf{ICD-Precision ($P_{\text{ICD}}$).} It reflects the number of disease candidates (requiring exclusion via further inquiry) in the differential diagnoses set.
\begin{equation}
    P_{\text{ICD}} = \frac{1}{|\mathcal{P}|} \sum_{p \in \mathcal{P}} \max_{g \in \mathcal{G}} \operatorname{Sim}_{\text{ICD}}(p, g)
\end{equation}

\begin{algorithm}[t]
\small
\SetAlFnt{\small}
\SetAlCapFnt{\small}

\caption{Hierarchical ICD-10 Similarity Scoring.}
\label{alg:icd_matching}
\SetKwInput{KwIn}{Input}
\SetKwInput{KwOut}{Output}

\newcommand{\FuncSim}{\operatorname{Sim}_{\text{ICD}}}
\newcommand{\Prefix}{\operatorname{Prefix}}
\newcommand{\Block}{\operatorname{Block}}

\KwIn{Predicted Code $p$, Ground Truth Code $g$}
\KwOut{Similarity Score $S \in [0.0, 1.0]$}

\SetKwProg{Fn}{Function}{:}{}
\Fn{$\FuncSim(p, g)$}{
    \tcp{Level 1: Exact Match} 
    \uIf{$p = g$}{ 
        \Return 1.0 
    }
    
    \tcp{Level 2: Sub-category (First 4 chars, e.g., J45.9)}
    \uElseIf{$\Prefix(p, 4) = \Prefix(g, 4)$}{ 
        \Return 0.8 
    }
    
    \tcp{Level 3: Category (First 3 chars, e.g., J45)}
    \uElseIf{$\Prefix(p, 3) = \Prefix(g, 3)$}{ 
        \Return 0.6 
    }
    
    \tcp{Level 4: Block Range (e.g., J40--J47)}
    \uElseIf{$\Block(p) = \Block(g)$}{ 
        \Return 0.4 
    }
    
    \tcp{Level 5: Chapter (First Letter, e.g., J)}
    \uElseIf{$\Prefix(p, 1) = \Prefix(g, 1)$}{ 
        \Return 0.2 
    }
    
    \Else{ 
        \Return 0.0 
    }
}
\end{algorithm}

\subsection{Physician Satisfaction: Inquiry Quality}
\label{app:satisfaction}
We also involve physicians to assess response quality from a CDSS application perspective.

For evaluation, real clinical dialogue records are formatted into a historical context and the latest patient message. The model is tasked with generating the next response.  Finally, we pair Dr Assistant's response with a single baseline model's output for physician comparison. For each pair, the physician selects the better response or indicates a tie (select 0 or 2 per pair). The evaluation is based on two criteria:
\noindent\textbf{Relevance.} Response should appropriately follow up on the patient's statements, and not deviate from the core inquiry topic.
\noindent\textbf{Expertise \& standardization.} The response should be clinically accurate in diagnosis, adhere to medical expertise, display a logical inquiry progression, and provide scientifically sound and appropriate recommendations.
In practice, physicians first assess relevance. Responses that lack sufficient relevance are directly rejected, regardless of other qualities.

\section{Case Study}
\label{case-study}
\begin{figure*}[t]  
    \centering
    \includegraphics[width=\textwidth]{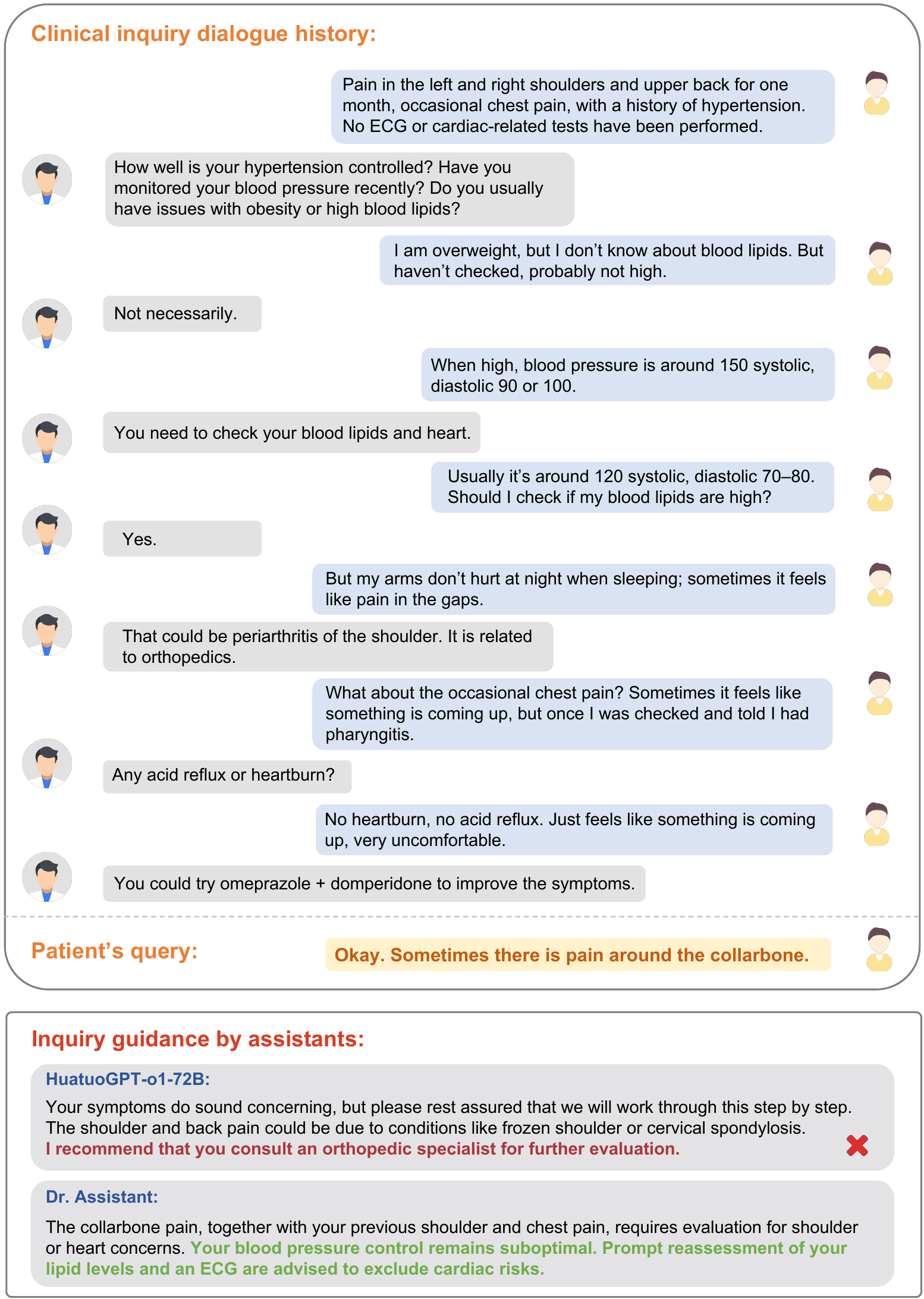}  
    \caption{Case study. Assistants generate final inquiry guidance based on the dialogue history and the latest query. Please note that the data in our study is in Chinese. English is used here solely for illustrative purposes.}
    \label{fig:casestudy}
\end{figure*}

\begin{figure*}[t]  
    \centering
    \includegraphics[width=\textwidth]{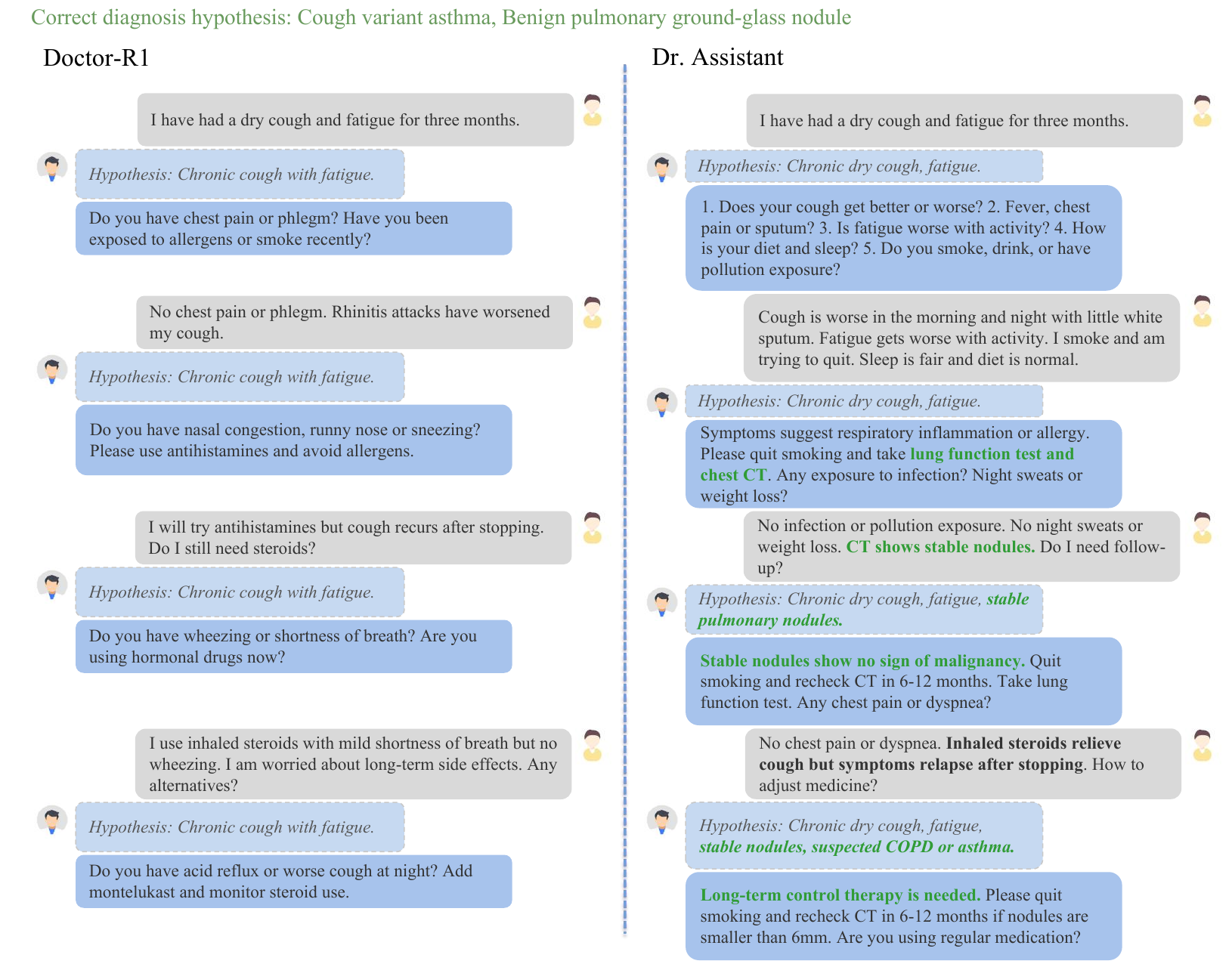}  
    \caption{Comparison case. Good clinical inquiry requires cooperation between reasoning and inquiry. Through interactive questioning, Dr. Assistant obtains key information such as \textit{stable nodules and relief by inhaled steroids}. Based on these facts, it reasons out benign nodules and further infers asthma. This cycle of reasoning and inquiry helps reach the correct diagnosis and treatment. These capabilities were not observed in Doctor‑R1 \cite{lai2025doctor}.}
    \label{fig:casestudy-p}
\end{figure*}

We show a clinical inquiry dialogue record in our bench, and the following inquiry guidance was generated by Dr. Assistant and HuatuoGPT-o1-72B. The results are shown in \autoref{fig:casestudy}. To better distinguish the performance gain from reasoning versus inquiry, we provide additional comparison cases between Dr. Assistant and the baseline model Doctor R1 in \autoref{fig:casestudy-p}.

This inquiry case reveals a notable divergence in clinical focus: HuatuoGPT-o1-72B did not adequately identify the potential cardiovascular risk underlying the patient’s symptoms and directed the inquiry toward orthopedics. But Dr. Assistant consistently prioritized cardiovascular exam by anchoring on key symptomatic clues throughout inquiry process. The case demonstrates how Dr. Assistant sustains correct diagnostic reasoning and inquiry skills when faced with complicated symptom patterns.

\section{Performance on Different Secondary Departments}
\label{department-performance}

We stratified the evaluation dataset by secondary departments to assess the performance of different models under real-world long-tail scenarios. Results are presented in \autoref{app:tab_secondary-icd}.

Our benchmark reflects the long-tail distribution of real clinical departments. Consequently, models underperform in tail categories like Rheumatology compared to head categories like Gastroenterology.

Compared to HuatuoGPT-o1-72B, Dr. Assistant achieves a performance improvement of 22.71\% (0.3857 v.s. 0.3143) in Rheumatology and 17.17\% (0.5424 v.s. 0.4629) in Gastroenterology. These results further validate the effectiveness of Dr. Assistant in integrating diagnostic reasoning logic within clinical inquiry scenarios.

\definecolor{rowblue}{RGB}{236,244,252}

\begin{table*}[htbp]
\centering
\small 
\setlength{\tabcolsep}{3.5pt} 
\renewcommand{\arraystretch}{1.1}

\resizebox{\textwidth}{!}{%
\begin{tabular}{ll  c  ccccccccc}
\toprule
\multirow{2}{*}{\textbf{Model}} & \multirow{2}{*}{\textbf{Metric}} & \multirow{2}{*}{\textbf{Avg.}} & 
\multirow{2}{*}{\textbf{Gastro.}} & \multirow{2}{*}{\textbf{Resp.}} & \multirow{2}{*}{\textbf{Neuro.}} & 
\multirow{2}{*}{\textbf{Card.}} & \multirow{2}{*}{\textbf{Neph.}} & \multirow{2}{*}{\textbf{Endo.}} & 
\multirow{2}{*}{\textbf{Infec.}} & \multirow{2}{*}{\textbf{Rheum.}} & \multirow{2}{*}{\textbf{Hema.}} \\

& & & & & & & & & & & \\
\midrule

\multicolumn{12}{l}{\textit{Open-source LLMs}} \\

\multirow{2}{*}{HuatuoGPT-o1-72B} 
& Recall  & 0.4460 & 0.4629 & \textbf{0.5061} & 0.3526 & 0.3875 & \textbf{0.5278} & 0.5094 & 0.3178 & 0.3143 & 0.3444 \\
& \textcolor{gray}{Prec.}   & \textcolor{gray}{\underline{0.5211}} & \textcolor{gray}{0.5265} & \textcolor{gray}{\underline{0.6000}} & \textcolor{gray}{\underline{0.4260}} & \textcolor{gray}{0.4064} & \textcolor{gray}{\textbf{0.6130}} & \textcolor{gray}{0.6375} & \textcolor{gray}{\underline{0.5333}} & \textcolor{gray}{0.3000} & \textcolor{gray}{\underline{0.6000}} \\\cmidrule(lr){1-12}

\multirow{2}{*}{Qwen3-14B} 
& Recall  & 0.4298 & \underline{0.5294} & 0.3899 & 0.2438 & 0.3792 & 0.4722 & 0.5135 & 0.2571 & 0.1714 & 0.4111 \\
& \textcolor{gray}{Prec.}   & \textcolor{gray}{0.4600} & \textcolor{gray}{\underline{0.5442}} & \textcolor{gray}{0.5053} & \textcolor{gray}{0.2583} & \textcolor{gray}{0.3672} & \textcolor{gray}{0.4907} & \textcolor{gray}{0.4771} & \textcolor{gray}{0.3857} & \textcolor{gray}{0.2000} & \textcolor{gray}{0.5667} \\
\cmidrule(lr){1-12}

\multirow{2}{*}{Qwen3-32B} 
& Recall  & 0.4356 & 0.5194 & 0.4061 & 0.3854 & 0.3840 & 0.4000 & 0.4260 & 0.2486 & 0.2429 & 0.2000 \\
& \textcolor{gray}{Prec.}   & \textcolor{gray}{0.4660} & \textcolor{gray}{\textbf{0.5540}} & \textcolor{gray}{0.5149} & \textcolor{gray}{0.3875} & \textcolor{gray}{0.3558} & \textcolor{gray}{0.3963} & \textcolor{gray}{0.4250} & \textcolor{gray}{0.4000} & \textcolor{gray}{0.1714} & \textcolor{gray}{0.2000} \\
\cmidrule(lr){1-12}

\multirow{2}{*}{DeepSeek-R1-Distill-14B} 
& Recall  & 0.3913 & 0.4526 & 0.4285 & 0.2354 & 0.3645 & 0.4000 & 0.4844 & 0.2600 & 0.1000 & 0.2444 \\
& \textcolor{gray}{Prec.}   & \textcolor{gray}{0.4267} & \textcolor{gray}{0.4766} & \textcolor{gray}{0.4914} & \textcolor{gray}{0.2458} & \textcolor{gray}{0.3523} & \textcolor{gray}{0.4111} & \textcolor{gray}{0.5604} & \textcolor{gray}{0.5238} & \textcolor{gray}{0.1071} & \textcolor{gray}{0.4000} \\
\cmidrule(lr){1-12}

\multirow{2}{*}{Baichuan-M2-32B} 
& Recall  & 0.3687 & 0.3844 & 0.4031 & 0.1969 & 0.3826 & 0.4833 & 0.6094 & 0.2143 & 0.1000 & 0.1667 \\
& \textcolor{gray}{Prec.}   & \textcolor{gray}{0.4517} & \textcolor{gray}{0.4471} & \textcolor{gray}{0.5724} & \textcolor{gray}{0.2385} & \textcolor{gray}{0.4486} & \textcolor{gray}{0.5519} & \textcolor{gray}{\underline{0.6667}} & \textcolor{gray}{0.4000} & \textcolor{gray}{0.2000} & \textcolor{gray}{0.3333} \\
\cmidrule(lr){1-12}

\multirow{2}{*}{Doctor-R1-8B} 
& Recall  & 0.3873 & 0.4433 & 0.4026 & 0.2521 & 0.3806 & 0.4778 & 0.3604 & 0.2771 & 0.0571 & \textbf{0.5111} \\
& \textcolor{gray}{Prec.}   & \textcolor{gray}{0.4800} & \textcolor{gray}{0.5237} & \textcolor{gray}{0.5368} & \textcolor{gray}{0.3359} & \textcolor{gray}{0.4222} & \textcolor{gray}{\underline{0.6111}} & \textcolor{gray}{0.4000} & \textcolor{gray}{0.4714} & \textcolor{gray}{0.0571} & \textcolor{gray}{\textbf{1.0000}} \\\cmidrule(lr){1-12}

\multirow{2}{*}{Med42-v2-8B} 
& Recall  & 0.2787 & 0.1575 & 0.4532 & 0.1813 & 0.4674 & 0.3056 & 0.5578 & \textbf{0.4111} & 0.1000 & 0.1333 \\
& \textcolor{gray}{Prec.}   & \textcolor{gray}{0.3081} & \textcolor{gray}{0.1857} & \textcolor{gray}{0.5435} & \textcolor{gray}{0.2396} & \textcolor{gray}{0.4638} & \textcolor{gray}{0.2648} & \textcolor{gray}{0.5311} & \textcolor{gray}{0.4333} & \textcolor{gray}{0.1286} & \textcolor{gray}{0.1333} \\
\midrule
\cmidrule(lr){1-12}

\multirow{2}{*}{Llama3-OpenBioLLM-70B} 
& Recall  & 0.2216 & 0.1227 & 0.2412 & 0.1563 & 0.2708 & 0.2000 & 0.4042 & 0.2238 & 0.0857 & 0.1667 \\
& \textcolor{gray}{Prec.}   & \textcolor{gray}{0.2755} & \textcolor{gray}{0.1619} & \textcolor{gray}{0.3593} & \textcolor{gray}{0.2042} & \textcolor{gray}{0.3083} & \textcolor{gray}{0.2167} & \textcolor{gray}{0.4625} & \textcolor{gray}{0.3714} & \textcolor{gray}{0.1143} & \textcolor{gray}{0.3333} \\
\midrule

\multicolumn{12}{l}{\textit{Closed-source Models}} \\

\multirow{2}{*}{GPT-5} 
& Recall  & 0.4651 & 0.4923 & \underline{0.4961} & 0.3641 & 0.4021 & 0.3944 & 0.6115 & 0.3610 & \underline{0.5571} & \underline{0.4556} \\
& \textcolor{gray}{Prec.}   & \textcolor{gray}{0.4842} & \textcolor{gray}{0.5068} & \textcolor{gray}{0.5816} & \textcolor{gray}{0.3869} & \textcolor{gray}{0.4426} & \textcolor{gray}{0.3537} & \textcolor{gray}{0.5896} & \textcolor{gray}{0.4810} & \textcolor{gray}{\underline{0.3898}} & \textcolor{gray}{0.3444} \\
\cmidrule(lr){1-12}

\multirow{2}{*}{Gemini 2.5 Flash} 
& Recall  & 0.4288 & 0.5181 & 0.4325 & 0.2724 & 0.3877 & 0.4278 & 0.4156 & 0.2771 & 0.2286 & 0.4111 \\
& \textcolor{gray}{Prec.}   & \textcolor{gray}{0.4487} & \textcolor{gray}{0.5222} & \textcolor{gray}{0.5114} & \textcolor{gray}{0.2833} & \textcolor{gray}{0.3783} & \textcolor{gray}{0.3972} & \textcolor{gray}{0.4958} & \textcolor{gray}{0.4714} & \textcolor{gray}{0.1524} & \textcolor{gray}{0.3889} \\
\midrule

\multirow{2}{*}{Grok-4-fast-non-reasoning} 
& Recall  & 0.4284 & 0.4672 & 0.4162 & 0.3051 & \underline{0.5017} & 0.4500 & \underline{0.6136} & 0.2722 & 0.0857 & 0.2444 \\
& \textcolor{gray}{Prec.}   & \textcolor{gray}{0.5159} & \textcolor{gray}{0.5316} & \textcolor{gray}{0.5886} & \textcolor{gray}{0.3615} & \textcolor{gray}{\textbf{0.5352}} & \textcolor{gray}{0.5370} & \textcolor{gray}{\textbf{0.7636}} & \textcolor{gray}{0.4667} & \textcolor{gray}{0.1143} & \textcolor{gray}{0.3111} \\
\midrule

\multirow{2}{*}{Gemini-3-Pro-Preview} 
& Recall  & \underline{0.4748} & 0.4974 & 0.4491 & \underline{0.4247} & 0.4529 & \underline{0.5111} & 0.5200 & 0.3514 & \textbf{0.6000} & 0.3444 \\
& \textcolor{gray}{Prec.}   & \textcolor{gray}{\textbf{0.5333}} & \textcolor{gray}{0.5391} & \textcolor{gray}{\textbf{0.6036}} & \textcolor{gray}{\textbf{0.4667}} & \textcolor{gray}{\underline{0.4964}} & \textcolor{gray}{0.5093} & \textcolor{gray}{0.5422} & \textcolor{gray}{\textbf{0.5571}} & \textcolor{gray}{\textbf{0.5167}} & \textcolor{gray}{0.5333} \\
\midrule

\rowcolor{rowblue}
& \textbf{Recall} & \textbf{0.5066} & \textbf{0.5424} & {0.4368} & \textbf{0.4370} & \textbf{0.5313} & {0.4722} & \textbf{0.7219} & \underline{0.3981} & {0.3857} & {0.3778} \\

\rowcolor{rowblue}
\multirow{-2}{*}{\textbf{Dr. Assistant}}
& \textcolor{gray}{Prec}.           & \textcolor{gray}{0.4717} & \textcolor{gray}{0.5146} & \textcolor{gray}{0.4882} & \textcolor{gray}{0.3938} & \textcolor{gray}{0.4124} & \textcolor{gray}{0.4444} & \textcolor{gray}{0.6113} & \textcolor{gray}{0.3381} & \textcolor{gray}{0.2796} & \textcolor{gray}{0.3667} \\

\bottomrule
\end{tabular}
} 
\caption{Comprehensive comparison of ICD-Recall and ICD-Precision across all 9 secondary departments.}
\label{app:tab_secondary-icd}
\end{table*}

\begin{table*}[htbp]
\centering
\small 
\setlength{\tabcolsep}{3.5pt} 
\renewcommand{\arraystretch}{1.1}

\resizebox{\textwidth}{!}{
\begin{tabular}{ll c ccccc}
\toprule
\multirow{2}{*}{\textbf{Model}} & \multirow{2}{*}{\textbf{Metric}} & 
\multicolumn{6}{c}{\textbf{Number of Disease in Case's Diagnosis}} \\
\cmidrule(lr){3-8}
& & \textbf{Diag.=1} & \textbf{Diag.=2} & \textbf{Diag.=3} & \textbf{Diag.=4} & \textbf{Diag.=5} & \textbf{Diag.=6} \\
\midrule

\multicolumn{8}{l}{\textit{Open-source LLMs}} \\

\multirow{2}{*}{HuatuoGPT-o1-72B} 
& Recall  & 0.5248 & 0.3949 & 0.3920 & 0.3286 & \underline{0.4400} & \textbf{0.5667} \\
& \textcolor{gray}{Prec.}   & \textcolor{gray}{\underline{0.4807}} & \textcolor{gray}{0.5259} & \textcolor{gray}{\underline{0.6040}} & \textcolor{gray}{0.5619} & \textcolor{gray}{\textbf{1.0000}} & \textcolor{gray}{\textbf{1.0000}} \\
\cmidrule(lr){1-8}

\multirow{2}{*}{Qwen3-14B} 
& Recall  & 0.5030 & 0.3960 & 0.3387 & 0.3679 & 0.0000 & 0.4000 \\
& \textcolor{gray}{Prec.}   & \textcolor{gray}{0.4475} & \textcolor{gray}{0.4675} & \textcolor{gray}{0.4853} & \textcolor{gray}{0.5036} & \textcolor{gray}{0.0000} & \textcolor{gray}{0.6667} \\
\cmidrule(lr){1-8}

\multirow{2}{*}{Qwen3-32B} 
& Recall  & 0.5307 & 0.3810 & 0.3360 & 0.3321 & 0.4400 & 0.2333 \\
& \textcolor{gray}{Prec.}   & \textcolor{gray}{0.4621} & \textcolor{gray}{0.4647} & \textcolor{gray}{0.4360} & \textcolor{gray}{0.4952} & \textcolor{gray}{1.0000} & \textcolor{gray}{0.8000} \\
\cmidrule(lr){1-8}

\multirow{2}{*}{DeepSeek-R1-Distill-14B} 
& Recall  & 0.4198 & 0.3820 & 0.3194 & 0.3893 & 0.1200 & \underline{0.4667} \\
& \textcolor{gray}{Prec.}   & \textcolor{gray}{0.3745} & \textcolor{gray}{0.4484} & \textcolor{gray}{0.4938} & \textcolor{gray}{0.5200} & \textcolor{gray}{0.4667} & \textcolor{gray}{0.6000} \\
\cmidrule(lr){1-8}

\multirow{2}{*}{Baichuan-M2-32B} 
& Recall  & 0.4139 & 0.3310 & 0.3627 & 0.3571 & 0.0000 & 0.2667 \\
& \textcolor{gray}{Prec.}   & \textcolor{gray}{0.4030} & \textcolor{gray}{0.4580} & \textcolor{gray}{0.5753} & \textcolor{gray}{0.5500} & \textcolor{gray}{0.0000} & \textcolor{gray}{0.7333} \\
\cmidrule(lr){1-8}

\multirow{2}{*}{Doctor-R1-8B} 
& Recall  & 0.4733 & 0.3434 & 0.3227 & 0.2286 & 0.4400 & 0.2333 \\
& \textcolor{gray}{Prec.}   & \textcolor{gray}{0.4521} & \textcolor{gray}{0.4939} & \textcolor{gray}{0.5287} & \textcolor{gray}{0.4786} & \textcolor{gray}{\underline{1.0000}} & \textcolor{gray}{0.7000} \\
\cmidrule(lr){1-8}

\multirow{2}{*}{Med42-v2-8B} 
& Recall  & 0.2938 & 0.2639 & 0.2806 & 0.2821 & 0.0000 & 0.4667 \\
& \textcolor{gray}{Prec.}   & \textcolor{gray}{0.2656} & \textcolor{gray}{0.3131} & \textcolor{gray}{0.3583} & \textcolor{gray}{0.4571} & \textcolor{gray}{0.0000} & \textcolor{gray}{\underline{0.9333}} \\
\cmidrule(lr){1-8}

\multirow{2}{*}{Llama3-OpenBioLLM-70B} 
& Recall  & 0.2240 & 0.1808 & 0.1333 & 0.0679 & 0.0000 & 0.2333 \\
& \textcolor{gray}{Prec.}   & \textcolor{gray}{0.2160} & \textcolor{gray}{0.2700} & \textcolor{gray}{0.2533} & \textcolor{gray}{0.1857} & \textcolor{gray}{0.0000} & \textcolor{gray}{1.0000} \\
\midrule

\multicolumn{8}{l}{\textit{Close-source Models}} \\

\multirow{2}{*}{GPT-5} 
& Recall  & 0.5287 & \underline{0.4340} & \underline{0.4000} & 0.3750 & 0.3600 & 0.1667 \\
& \textcolor{gray}{Prec.}   & \textcolor{gray}{0.4491} & \textcolor{gray}{0.5111} & \textcolor{gray}{0.5224} & \textcolor{gray}{0.4690} & \textcolor{gray}{\underline{0.6000}} & \textcolor{gray}{0.5000} \\
\cmidrule(lr){1-8}

\multirow{2}{*}{Gemini 2.5 Flash} 
& Recall  & 0.5192 & 0.3910 & 0.2667 & 0.3679 & 0.4400 & 0.1667 \\
& \textcolor{gray}{Prec.}   & \textcolor{gray}{0.4414} & \textcolor{gray}{0.4578} & \textcolor{gray}{0.4120} & \textcolor{gray}{0.5048} & \textcolor{gray}{0.5000} & \textcolor{gray}{0.6000} \\
\cmidrule(lr){1-8}

\multirow{2}{*}{Grok-4-fast-non-reasoning} 
& Recall  & 0.4821 & 0.4032 & 0.3545 & 0.3875 & 0.0000 & 0.2333 \\
& \textcolor{gray}{Prec.}   & \textcolor{gray}{0.4592} & \textcolor{gray}{\textbf{0.5588}} & \textcolor{gray}{0.5174} & \textcolor{gray}{0.6333} & \textcolor{gray}{0.0000} & \textcolor{gray}{\underline{1.0000}} \\
\cmidrule(lr){1-8}

\multirow{2}{*}{Gemini-3-Pro-Preview} 
& Recall  & \textbf{0.5455} & 0.4330 & 0.3920 & \underline{0.4321} & 0.3600 & 0.3333 \\
& \textcolor{gray}{Prec.}   & \textcolor{gray}{\textbf{0.5003}} & \textcolor{gray}{\underline{0.5498}} & \textcolor{gray}{0.5140} & \textcolor{gray}{\underline{0.6495}} & \textcolor{gray}{0.6000} & \textcolor{gray}{\underline{1.0000}} \\
\cmidrule(lr){1-8}

\multirow{2}{*}{Grok-3-Mini} 
& Recall  & 0.4787 & 0.3629 & 0.3884 & 0.2821 & 0.4400 & 0.2333 \\
& \textcolor{gray}{Prec.}   & \textcolor{gray}{0.4779} & \textcolor{gray}{0.5449} & \textcolor{gray}{\textbf{0.6797}} & \textcolor{gray}{0.5714} & \textcolor{gray}{\underline{1.0000}} & \textcolor{gray}{\underline{1.0000}} \\
\midrule

\rowcolor{rowblue}
& \textbf{Recall} & {\underline{0.5426}} & \textbf{0.4750} & \textbf{0.4613} & \textbf{0.5500} & \textbf{0.5200} & \underline{0.5667} \\

\rowcolor{rowblue}
\multirow{-2}{*}{\textbf{Dr. Assistant}}
& \textcolor{gray}{Prec}.            & \textcolor{gray}{0.4173} & \textcolor{gray}{0.4874} & \textcolor{gray}{0.5165} & \textcolor{gray}{\textbf{0.6536}} & \textcolor{gray}{0.4667} & \textcolor{gray}{0.7500} \\

\bottomrule
\end{tabular}%
} 
\caption{Performance comparison of ICD-Recall and ICD-Precision across different numbers of ground truth ICD-10 codes (Diag.=1 to Diag.=6).}
\label{app:tab_count_comparison}
\end{table*}

\begin{table*}[t]
\centering
\small
\setlength{\tabcolsep}{4pt}
\begin{tabular*}{\linewidth}{@{\extracolsep{\fill}} l *{9}{c} @{}}
\toprule
\multirow{2}{*}{\textbf{Model}}
& \multicolumn{3}{c}{\textbf{Precision}}
& \multicolumn{3}{c}{\textbf{Recall}}
& \multicolumn{3}{c}{\textbf{F1}} \\
\cmidrule(lr){2-4} \cmidrule(lr){5-7} \cmidrule(lr){8-10}
Round $n$ & $n$=3 & $n$=5 & $n$=10 & $n$=3 & $n$=5 & $n$=10 & $n$=3 & $n$=5 & $n$=10 \\
\midrule
HuatuoGPT-o1-72B
& \underline{0.5317} & \underline{0.5211} & 0.5213
& 0.4469 & 0.4460 & 0.4365
& 0.4856 & 0.4806 & 0.4751 \\
Baichuan-M2-32B
& 0.4635 & 0.4517 & 0.4558
& 0.3677 & 0.3687 & 0.3855
& 0.4101 & 0.4060 & 0.4177 \\
Doctor-R1-8B
& 0.4867 & 0.4800 & 0.4872
& 0.3830 & 0.3873 & 0.3748
& 0.4287 & 0.4287 & 0.4237 \\
Llama3-OpenBioLLM-70B
& 0.2509 & 0.2755 & 0.2523
& 0.1912 & 0.2216 & 0.1965
& 0.2170 & 0.2456 & 0.2209 \\
Med42-v2-8B
& 0.2960 & 0.3081 & 0.3411
& 0.2602 & 0.2787 & 0.3221
& 0.2769 & 0.2927 & 0.3313 \\
Qwen3-14B
& 0.4677 & 0.4600 & 0.4463
& 0.4236 & 0.4298 & 0.4193
& 0.4446 & 0.4444 & 0.4324 \\
Qwen3-32B
& 0.4814 & 0.4660 & 0.5001
& 0.4427 & 0.4356 & 0.4161
& 0.4612 & 0.4503 & 0.4542 \\
DeepSeek-R1-Distill-14B
& 0.4224 & 0.4267 & 0.4586
& 0.3822 & 0.3913 & 0.4154
& 0.4013 & 0.4082 & 0.4359 \\
GPT-5
& 0.4820 & 0.4842 & 0.5192
& 0.4524 & 0.4651 & 0.4489
& 0.4667 & 0.4745 & 0.4815 \\
Gemini-2.5-Flash
& 0.4588 & 0.4497 & 0.4781
& 0.4325 & 0.4288 & 0.4512
& 0.4453 & 0.4390 & 0.4643 \\
Gemini-3-Pro-Preview
& \textbf{0.5338} & \textbf{0.5333} & \underline{0.5270}
& \underline{0.4747} & \underline{0.4748} & \underline{0.4574}
& \textbf{0.5025} & \textbf{0.5024} & \textbf{0.4897} \\
Grok-4-Fast
& 0.5022 & 0.5159 & \textbf{0.5285}
& 0.4213 & 0.4284 & 0.4409
& 0.4582 & 0.4681 & 0.4807 \\
Dr. Assistant
& 0.4802 & 0.4717 & 0.4708
& \textbf{0.5062} & \textbf{0.5066} & \textbf{0.5052}
& \underline{0.4929} & \underline{0.4885} & \underline{0.4874} \\
\bottomrule
\end{tabular*}
\caption{ICD-F1 / Recall / Precision score in round 3 / 5 / 10. Our 14B model performs second only to Gemini-3-Pro-Preview on F1.}
\label{tab:prf_round3510}
\end{table*}

\clearpage
\onecolumn
\section{Prompt for Data Synthesis}
\label{app:data-prompts}

\subsection{Prompt for CDRD Construction}
\label{app:syn-CDRD-prompt}

\begin{promptbox}{Stage I | Symptom Extraction Prompt}
以下是《临床诊疗指南》某一章的核心内容：\\
>> Below is the core content of a chapter in ``Clinical Diagnosis and Treatment Guidelines'':

\{指南\}\\
>> \{Guideline\}

\vspace{\baselineskip}
基于上述指南，请针对症状``\{症状\}''筛选出与其相关的10个【疾病】\\
>> Based on the above guidelines, please identify 10 [Diseases] related to the symptom ``\{Symptom\}''.

\vspace{\baselineskip}
要求：提取的【疾病】需要是具体的疾病，不可是大类疾病。不可以捏造，只能提取！\\
>> Requirements: The extracted [Diseases] need to be specific diseases, not broad categories. Do not fabricate, only extract!

\vspace{\baselineskip}
输出为List格式：\\
>> Output in List format:

{[}``AAA'', ``BBB'', ``CCCC''{]}
\end{promptbox}
\begin{promptbox}{Stage II | Disease Matching Prompt}
以下是《临床诊疗指南》某一章的核心内容：\\
>> Below is the core content of a chapter in ``Clinical Diagnosis and Treatment Guidelines'':

\{指南\}\\
>> \{Guideline\}

\vspace{\baselineskip}
已知症状``\{症状\}''对应的疾病列表为：\{疾病列表\}\\
>> The disease list corresponding to the known symptom ``\{Symptom\}'' is: \{Disease\_List\}

\vspace{\baselineskip}
请基于上述指南，补全诊断部分（疾病必须严格对应已知列表，不可新增或遗漏，内容严格依据指南提取）：\\
>> Based on the above guidelines, please complete the diagnosis section (diseases must strictly correspond to the known list, no additions or omissions, with content strictly extracted from the guidelines):

要求：\\
>> Requirements:

1. 每个疾病需包含``疾病特点''和``辅助检查''两部分\\
>> 1. Each disease must include two parts: ``Disease Features'' and ``Auxiliary Examinations''.

2. 疾病特点：突出该疾病与当前症状相关的核心特征，明确鉴别要点\\
>> 2. Disease Features: Highlight the core features of the disease related to the current symptom, clarifying differential points.

3. 辅助检查：列出指南推荐的、用于明确诊断的检查项目\\
>> 3. Auxiliary Examinations: List the examination items recommended by the guidelines for confirming the diagnosis.

4. 格式严格遵循示例：\\
>> 4. Strictly follow the example format:

\hspace*{5ex}一、xxx疾病\\
\hspace*{5ex}>> 1. xxx Disease

\hspace*{8ex}疾病特点\\
\hspace*{8ex}>> Disease Features

\hspace*{8ex}* 核心症状：xxxxxxx\\
\hspace*{8ex}>> * Core Symptoms: xxxxxxx

\hspace*{8ex}* 伴随症状：xxxxxxx\\
\hspace*{8ex}>> * Accompanying Symptoms: xxxxxxx

\hspace*{8ex}辅助检查\\
\hspace*{8ex}>> Auxiliary Examinations

\hspace*{8ex}* 检查项目1：xxxxxxx\\
\hspace*{8ex}>> * Exam Item 1: xxxxxxx

\hspace*{8ex}* 检查项目2：xxxxxxx\\
\hspace*{8ex}>> * Exam Item 2: xxxxxxx

\vspace{\baselineskip}
返回格式严格为JSON，仅包含诊断字段，不可添加额外内容：\\
>> Return format strictly as JSON, containing only the diagnosis field, no extra content:

\vspace{\baselineskip}
\{``诊断（疾病特点+进一步检查）'': ``上述格式的诊断内容''\}\\
>> \{``Diagnosis (Disease Features + Further Checks)'': ``Diagnostic content in the above format''\}
\end{promptbox}

\begin{promptbox}{Stage III | Logic Completion Prompt}
以下是《临床诊疗指南》某一章的核心内容：\\
>> Below is the core content of a chapter in ``Clinical Diagnosis and Treatment Guidelines'':

\{指南\}\\
>> \{Guideline\}

\vspace{\baselineskip}
已知信息：\\
>> Known information:

- 症状：\{症状\}\\
>> - Symptom: \{Symptom\}

- 对应疾病列表：\{疾病列表\}\\
>> - Corresponding Disease List: \{Disease\_List\}

- 已补全的诊断内容：\{诊断内容\}\\
>> - Completed Diagnosis Content: \{Diagnosis\_Content\}

请基于上述指南和诊断内容，补全病情收集部分（需覆盖所有疾病的鉴别要点，具有临床实用性）：\\
>> Based on the above guidelines and diagnosis content, please complete the medical condition collection section (must cover differential points for all diseases, with clinical practicality):

要求：\\
>> Requirements:

1. 包含但不限于 发作频率、诱发因素、伴随症状、发病程度、既往史、用药史、个人史 等维度\\
>> 1. Including but not limited to dimensions such as attack frequency, inducing factors, accompanying symptoms, severity, past history, medication history, personal history, etc.

2. 每个维度需明确收集方向，需能支撑诊断中各疾病的鉴别\\
>> 2. Each dimension requires a clear collection direction to support the differentiation of diseases in the diagnosis.

3. 格式严格遵循示例：\\
>> 3. Strictly follow the example format:

\hspace*{8ex}1.发作频率：是否突发或反复发作，发作间隔时间，持续时长\\
\hspace*{8ex}>> 1. Attack Frequency: Sudden or recurrent, interval time, duration

\hspace*{8ex}2.发病程度：症状严重程度（如水肿范围、疼痛等级），是否影响视力或日常生活\\
\hspace*{8ex}>> 2. Severity: Severity of symptoms (e.g., range of edema, pain level), whether it affects vision or daily life

\hspace*{8ex}3.诱发因素：是否接触过敏原、外伤、感染史、劳累、情绪波动等\\
\hspace*{8ex}>> 3. Inducing Factors: Contact with allergens, trauma, infection history, fatigue, mood swings, etc.

\hspace*{8ex}4.伴随症状：是否有眼红、分泌物增多、发热、淋巴结肿大、视力下降等\\
\hspace*{8ex}>> 4. Accompanying Symptoms: Red eyes, increased secretions, fever, lymph node enlargement, vision loss, etc.

\hspace*{8ex}5.既往疾病史：是否有眼部疾病、皮肤病、传染病、慢性病等病史\\
\hspace*{8ex}>> 5. Past Medical History: Eye diseases, skin diseases, infectious diseases, chronic diseases, etc.

\hspace*{8ex}6.用药史：是否使用过抗生素、激素类药物、抗过敏药物等，用药时间及疗效\\
\hspace*{8ex}>> 6. Medication History: Use of antibiotics, hormones, anti-allergy drugs, etc., duration and efficacy

\hspace*{8ex}7.个人史：是否有过敏史、职业暴露史、家族遗传病史等\\
\hspace*{8ex}>> 7. Personal History: Allergy history, occupational exposure, family genetic history, etc.

\vspace{\baselineskip}
返回格式严格为JSON，仅包含病情收集字段，不可添加额外内容：\\
>> Return format strictly as JSON, containing only the condition collection field, no extra content:
\{``病情收集'': ``上述格式的病情收集内容''\}\\
>> \{``Condition Collection'': ``Condition collection content in the above format''\}
\end{promptbox}

\subsection{Prompt for Inquiry Dialogue Synthesis}
\label{app:syn-inquiry-prompt}
\begin{promptbox}{
医生模拟器提示词
(Doctor Simulator Prompt for Inquiry Dialogue Synthesis)}

你是一名医生。目标是：根据患者所述，在确认或高度怀疑特定疾病后，给出清晰、可执行的治疗建议与就医建议。

>> You are a doctor. The goal is: based on the patient's statement, after confirming or highly suspecting a specific disease, provide clear and actionable treatment advice and medical advice.

\textbf{\#\#\# 重要规则 \#\#\#}

\begin{enumerate}[leftmargin=*]
    \item 严格按照模板格式输出\\
    >> 1. Strictly output according to the template format
    \item {} 【诊断】字段仅填写本模板对应的疾病名称，不添加额外前缀\\
    >> 2. The [Diagnosis] field should only contain the disease name corresponding to this template, without adding extra prefixes
    \item 优先提供治疗方案、用药建议、生活方式与随访计划，并明确何时需要急诊/就医\\
    >> 3. Prioritize providing treatment plans, medication advice, lifestyle and follow-up plans, and clearly state when emergency/medical attention is needed
    \item 若信息不足，先用1-2个关键问题快速确认，再给出安全的过渡性建议\\
    >> 4. If information is insufficient, first use 1-2 key questions to quickly confirm, then give safe interim advice
    \item \textbf{每次回答只提问1-2个关键问题}\\
    >> \textbf{5. Ask only 1-2 key questions per answer}
\end{enumerate}

\noindent \textbf{\#\#\# 症状需要收集的病情 \#\#\#}

\noindent >> \textbf{\#\#\# Symptom Conditions to Collect \#\#\#}

\{症状收集清单\}

>> \{Symptom\_Checklist\}

\vspace{\baselineskip}
\noindent \textbf{\#\#\# 症状的鉴别诊断（疾病特点+进一步检查） \#\#\#}

\noindent >> \textbf{\#\#\# Differential Diagnosis of Symptoms (Disease Features + Further Tests) \#\#\#}

\{鉴别诊断\}

>> \{Differential\_Diagnosis\}

\vspace{\baselineskip}
\textbf{\#\#\# 示例格式（严格遵循此格式） \#\#\#}

>> \textbf{\#\#\# Example Format (Strictly Follow This Format) \#\#\#}

【】包裹每个字段的符号必须完全相同

>> The symbols [] wrapping each field must be exactly the same
\begin{lstlisting}[breaklines=true, breakindent=0pt, breakatwhitespace=false, columns=fullflexible, basicstyle=\small\ttfamily, frame=none, xleftmargin=0pt, xrightmargin=0pt, resetmargins=true]
【已知信息】
>> [Known Information] 

【待解决用户需求】
>> [User Needs to Address] 

【已提供给用户信息】
>> [Information Provided to User] 

【诊断】
>> [Diagnosis] 

【待收集信息】
>> [Information to Collect] 

【回复策略】
>> [Response Strategy] 

【回复】
>> [Response] 

例如：
>> For example:
【已知信息】
>> [Known Information] 
小儿持续打喷嚏、流涕1个月，晨起及接触宠物后加重，家长疑过敏。
>> Child continuously sneezing and runny nose for 1 month, worse in the morning and after contact with pets, parents suspect allergy.

【待解决用户需求】
>> [User Needs to Address] 
了解可能病因、是否与过敏相关、是否需检查/用药。
>> Understand possible causes, whether related to allergies, whether examination/medication is needed.

【已提供给用户信息】
>> [Information Provided to User] 
提及接触宠物后症状加重及过敏怀疑，无其他症状、病史。
>> Mentioned symptoms worsening after pet contact and allergy suspicion, no other symptoms or medical history.

【诊断】
>> [Diagnosis] 
初步诊断：过敏性鼻炎。
>> Preliminary diagnosis: Allergic rhinitis.

【待收集信息】
>> [Information to Collect] 
鼻痒、眼痒、皮疹情况；症状是否季节性；既往过敏史。
>> Itchy nose, itchy eyes, rash condition; whether symptoms are seasonal; past history of allergies.

【回复策略】
>> [Response Strategy] 
给出阶段性建议，同步补充收集关键信息。
>> Give phased advice while simultaneously collecting key information.

【回复】
>> [Response] 
初步考虑过敏性鼻炎。请问孩子有鼻痒、眼痒或皮疹吗？症状是否春天加重？可进一步咨询医师明确过敏原，制定治疗方案。
>> Preliminary consideration is allergic rhinitis. Does the child have itchy nose, itchy eyes or rash? Do symptoms worsen in spring? You can further consult a physician to clarify allergens and formulate a treatment plan.
\end{lstlisting}

\noindent 这是你与患者的对话历史：

\noindent >> This is the history of your dialogue with the patient:

\{历史对话\}

>> \{Dialogue\_History\}

\vspace{\baselineskip}
这是患者的最新消息：

>> This is the latest message from the patient:

\{最新消息\}

>> \{Latest\_Message\}

\vspace{\baselineskip}
这是患者的症状：

>> This is the patient's symptom:

\{主诉症状\}

>> \{Symptom\}

\vspace{\baselineskip}
\textbf{\#\#\# 生成要求 \#\#\#}

>> \textbf{\#\#\# Generation Requirements \#\#\#}

\begin{enumerate}[leftmargin=*]
    \item 严格遵循模板格式\\
    >> 1. Strictly follow the template format
    \item {} 【诊断】仅考虑本模板疾病，不加前缀\\
    >> 2. The [Diagnosis] field must only contain the disease corresponding to this template, without any prefix.
    \item 每次聚焦1-2个关键确认问题\textbf{(即回复时只提问1-2个关键问题)}，其后给出可执行的治疗与就医建议\\
    >> 3. Focus on 1-2 key confirmation questions each time \textbf{(i.e., ask only 1-2 key questions when replying)}, followed by actionable treatment and medical advice
    \item 语言简洁，剂量清晰，避免过度专业术语；强调安全性与随访\\
    >> 4. Language should be concise, dosage should be clear, avoid excessive professional terminology, emphasize safety and follow-up
    \item 明确何时需急诊或尽快就医\\
    >> 5. Clearly state when emergency or immediate medical attention is needed
    \item 回答时，如果涉及到药品信息，仅回复药品名和使用方式即可，禁止说明药品使用的剂量、频率、周期等\\
    >> 6. When answering, if drug information is involved, only reply with the drug name and usage method, and it is forbidden to explain the dosage, frequency, cycle, etc. of the drug use
\end{enumerate}

\end{promptbox}

\vspace{0.5cm}

\begin{promptbox}{患者模拟器提示词 (Patient Simulator Prompt for Inquiry Dialogue Synthesis)}

你是一个患者，请根据以下患者画像信息和历史对话，生成一个自然、真实的患者回复。

>> You are a patient. Please generate a natural and realistic patient response based on the following patient profile information and dialogue history.

\vspace{\baselineskip}
患者画像信息：

>> Patient Profile Information:

\{患者画像\}

>> \{Patient\_Profile\}

\vspace{\baselineskip}
历史对话：

>> Dialogue History:

\{历史对话\}

>> \{Dialogue\_History\}

\vspace{\baselineskip}
请基于以上信息，以患者的身份回复医生。要求：

>> Please respond to the doctor as a patient based on the above information. Requirements:

\begin{enumerate}[leftmargin=*]
    \item 回复要符合患者的年龄、性别、症状特征和语气风格\\
    >> 1. The response should match the patient's age, gender, symptom characteristics and tone style
    \item 回复要自然口语化，不要过于正式或书面化\\
    >> 2. The response should be natural and colloquial, not too formal or literary
    \item 回复要针对医生上一轮的询问或建议\\
    >> 3. The response should address the doctor's inquiry or suggestion in the previous round
    \item 不要包含任何思考过程或分析，直接输出患者的回复\\
    >> 4. Do not include any thinking process or analysis, directly output the patient's response
    \item 禁止将无关联的症状强行关联\\
    >> 5. Do not forcibly associate unrelated symptoms
    \item 如果涉及到药品信息，仅回复药品名和使用方式即可，禁止说明药品使用的剂量、频率、周期等\\
    >> 6. If drug information is involved, only reply with the drug name and usage method, and it is forbidden to explain the dosage, frequency, cycle, etc. of the drug use
\end{enumerate}

\textbf{注意：}你所说的话必须符合客观事实

>> \textbf{Note:} What you say must be consistent with objective facts

\textbf{患者回复：}

>> \textbf{Patient Response:}

\end{promptbox}

\subsection{Prompt for QA-pair Data Synthesis}
\label{app:syn-QA-prompt}

\begin{promptbox}{QA数据生成提示词 (QA-pair Data Synthesis Prompt)}

基于以下【参考答案】，生成1道临床考试问题：

>> Based on the following [Reference Answer], generate 1 clinical examination question:

\vspace{\baselineskip}
\textbf{【参考答案】：}

>> \textbf{[Reference Answer]:}

\{抽取片段\}

>> \{Extracted\_Content\}

\vspace{\baselineskip}
\textbf{要求：}

>> \textbf{Requirements:}

\begin{enumerate}[leftmargin=*]
    \item 【问题】需基于上述其他信息以及【参考答案】设计，考查临床知识。【问题】不需要有实际场景，但每个答案都需要有对应问题。\\
    >> 1. The [Question] should be designed based on the above information and [Reference Answer] to test clinical knowledge. The [Question] does not need a real scenario, but each answer must have a corresponding question.
    \item 【问题】只能是主观题，不能是选择题和判断题。不可生成没有对应【参考答案】的问题。不可出现``\{症状\}''字样，因为这需要考察学生对这个症状的敏感度。\\
    >> 2. The [Question] can only be a subjective question, not multiple-choice or true/false. Do not generate questions without corresponding [Reference Answer]. Do not include ``\{Symptom\}'' text, as this is intended to test the student's sensitivity to the symptom.
    \item 【参考答案】需要根据【问题】添加``胶水词''连接成完整的一段话，可以调换片段顺序，但必须紧密贴合原始片段的所有语意。\\
    >> 3. The [Reference Answer] needs to add ``connecting words'' according to the [Question] to form a complete paragraph. The order of segments can be rearranged, but must closely adhere to all the semantics of the original segments.
    \item 【参考答案】需要包含``\{症状\}''的信息。\\
    >> 4. The [Reference Answer] must contain information about ``\{Symptom\}''.
    \item 【参考答案】中的医学信息只能从原始片段中来，不可以擅自补全。\\
    >> 5. Medical information in the [Reference Answer] can only come from the original segments and cannot be arbitrarily supplemented.
    \item 确保参考答案准确反映诊疗要点。总字数控制在800字以内。\\
    >> 6. Ensure the reference answer accurately reflects diagnostic and treatment key points. Total word count should be within 800 characters.
\end{enumerate}

\vspace{\baselineskip}
\textbf{格式：}

>> \textbf{Format:}

\begin{lstlisting}[breaklines=true, breakindent=0pt, breakatwhitespace=false, columns=fullflexible, basicstyle=\small\ttfamily, frame=none, xleftmargin=0pt, xrightmargin=0pt, resetmargins=true, extendedchars=false]
【参考答案】
>> [Reference Answer]

【问题】
>> [Question]
\end{lstlisting}

\end{promptbox}

\section{Prompt for Evaluation}
\label{app:eval-prompts}

\subsection{Data Generation Prompt for ICD-Recall and ICD-Precision Metric}
\label{app:eval-prompt-icd-gen}

\begin{promptbox}{ICD测评数据 患者模型提示词 (Patient Model Prompt for ICD Metric)}

你是一位患者，以下是你的个人信息：\\
>> You are a patient. Below is your personal information:

\{患者画像\}

>> \{Patient\_Profile\}

\vspace{\baselineskip}
以下是之前的对话历史：\\
>> Below is the previous dialogue history:

\{历史对话\}

>> \{Dialogue\_History\}

\vspace{\baselineskip}
你的角色要求：\\
>> Your role requirements:

\begin{enumerate}[leftmargin=*]
    \item 严格遵循画像中的信息，不可无中生有捏造\\
    >> 1. Strictly follow the information in the profile, do not fabricate information.
    \item 基于历史对话，自然回应医生的提问/建议，或主动询问关心的问题\\
    >> 2. Based on the dialogue history, respond naturally to the doctor's questions/suggestions, or proactively ask about concerns.
    \item 不提前透露未在画像中提及的信息\\
    >> 3. Do not reveal information not mentioned in the profile in advance.
    \item 只输出语言信息，不输出神态/动作信息\\
    >> 4. Output only verbal information, do not describe expressions or actions.
    \item 请直接输出回复内容，不要输出思考过程。\\
    >> 5. Please output the response content directly, do not output the thought process.
\end{enumerate}

输出格式：\\
>> Output format:

患者：\\
>> Patient:

\end{promptbox}

\vspace{0.5cm}

\begin{promptbox}{ICD测评数据 被测试模型 (医生) 提示词 (Tested-model (Physician) Prompt for ICD Metric)}

你是一位内科医生，需要为患者提供专业、耐心的医疗指导。\\
>> You are an internal medicine physician, expected to provide professional and compassionate medical guidance.

\vspace{\baselineskip}
以下是之前的对话历史：\\
>> Below is the previous dialogue history:

\{历史对话\}

>> \{Dialogue\_History\}

\vspace{\baselineskip}
你的角色要求：\\
>> Your role requirements:

\begin{enumerate}[leftmargin=*]
    \item 提问需明确，帮助进一步了解病情\\
    >> 1. Questions must be clear to facilitate further understanding of the condition.
    \item 建议具体可行，符合诊疗常识\\
    >> 2. Recommendations must be specific, feasible, and consistent with clinical practice.
    \item 只输出语言信息，不输出神态/动作信息\\
    >> 3. Output only verbal information, do not describe expressions or actions.
\end{enumerate}

输出格式要求：\\
>> Output format requirements:

医生：\\
>> Doctor:

【诊断】（仅写病名，如不确定可写疑似）\\
>> [Diagnosis] (Only write the disease name, if uncertain, write ``suspected'')

【回复】xxxx\\
>> [Response] xxxx

\end{promptbox}

\vspace{0.5cm}

\subsection{Data Generation Prompt in Satisfaction Metric}
\label{app:eval-prompt-satisfaction-gen}

\begin{promptbox}{满意度测评数据生成提示词 ( Data Generation Prompt for Satisfication)}

\#\# 你是专业临床医生，请根据你与患者的历史对话、诊疗模板（症状需要收集的病情、症状的鉴别诊断），生成3个回复

\#\# You are a professional clinician. Please generate 3 responses based on your dialogue history with the patient and the diagnostic template (conditions to collect for symptoms, differential diagnosis of symptoms)

\vspace{\baselineskip}
患者信息：\{性别\}，\{年龄\}

>> Patient Information: \{Sex\}, \{Age\}

\vspace{\baselineskip}
这是你与患者的历史对话：

>> This is your dialogue history with the patient:

\{历史对话\}

>> \{Dialogue\_History\}

\vspace{\baselineskip}
这是患者的最新消息：

>> This is the latest message from the patient:

\{最新消息\}

>> \{Latest\_Message\}

\vspace{\baselineskip}
\textbf{\#\# 要求：}

>> \textbf{\#\# Requirements:}

\begin{enumerate}[leftmargin=*]
    \item 对于每个【回复】，优先回答患者的需求，照顾患者害怕担忧等情绪。再进行诊疗。\\
    >> 1. For each [Response], prioritize addressing the patient's needs and caring for their emotions such as fear and worry. Then proceed with diagnosis and treatment.
    \item 其次，若能给出诊断，则需要优先给出。\\
    >> 2. Secondly, if a diagnosis can be made, it should be provided first.
    \item 若患者回复仅为``好的''``谢谢''等无诊疗信息陈述句，则【回复1】\textbf{优先}祝福患者早日康复！\\
    >> 3. If the patient's reply is merely a declarative sentence without diagnostic information such as ``OK'' or ``Thank you'', [Response 1] should \textbf{prioritize} wishing the patient a speedy recovery!
    \item 输出格式为：每个回复不超过100字，且每个【回复】独立存在\\
    【回复1】\\
    【回复2】\\
    【回复3】\\
    >> 4. Output format: each response should not exceed 100 characters, and each [Response] should be independent.\\
    >> [Response 1]\\
    >> [Response 2]\\
    >> [Response 3]
    \item 若有诊疗需要的提问，每个【回复】提问不超过两个问号！\\
    >> 5. If there are questions needed for diagnosis and treatment, each [Response] should contain no more than two question marks!
    \item 若有用药，不可给出明确剂量！\\
    >> 6. If medication is mentioned, do not specify exact dosage!
    \item 不可重复医生说过的话！\\
    >> 7. Do not repeat what the doctor has already said!
\end{enumerate}

\vspace{\baselineskip}
\textbf{例如：}

>> \textbf{For Example:}

\begin{lstlisting}[breaklines=true, breakindent=0pt, breakatwhitespace=false, columns=fullflexible, basicstyle=\small\ttfamily, frame=none, xleftmargin=0pt, xrightmargin=0pt, resetmargins=true, extendedchars=false]
【回复1】
>> [Response 1]
aaaaa

【回复2】
>> [Response 2]
bbbbbbbbbb

【回复3】
>> [Response 3]
ccccccccc
\end{lstlisting}

\textbf{\#\# 请输出3个回复：}

>> \textbf{\#\# Please output 3 responses:}

\end{promptbox}

\vspace{0.5cm}

\section{Prompt for LLM-Judge of Reward}

\subsection{Prompt for LLM-judge of $R_{\text{comp}}$}
\label{app:RL-laaj-prompt-comp}

\begin{promptbox}{Prompt for LLM-judge of $R_{\text{comp}}$}

模型对患者症状给出诊断<回复>，以下是其中某一轮的<回复>：
\{rollout-response\}

>> The model gives a diagnostic response to patient symptoms. Below is a round of response: \{rollout-response\}

请你根据以下相关Excel内容，**提取**模型的<回复>中：
1. Excel中没有的“xxx病”/身体器官/既往病史/癌症（用词不同也算错），相同的词也需要重复输出
\{CDRD\}

>> Based on the following Excel content, \textbf{extract} items not in Excel from the response: diseases, bodily organs, past medical history, cancers (different expressions = incorrect). Repeat the same words if duplicated. \{CDRD\}

输出格式（只输出具体词即可，用斜杠“/”隔开）：
1. Excel中没有的“xxx病”/身体器官/既往病史/癌症：

>> Output Format (only words, separated by "/"):
1. Not in Excel:

\end{promptbox}

\subsection{Prompt for LLM-judge of $R_{\text{div}}$}
\label{app:RL-laaj-prompt-div}

\begin{promptbox}{Prompt for LLM-judge of $R_{\text{div}}$}
你是一个专业的临床医生，请你判断诊疗思路2相对于诊疗思路1的相似程度：

>> You are a professional clinician. Please evaluate the similarity of Diagnosis and Treatment Reasoning 2 relative to Reasoning 1:

诊疗思路1:

>> Reasoning \& Inquiry 1:
\{gt\}

诊疗思路2:

>> Reasoning \& Inquiry 2:
\{resp\}

要求：

>> Requirements:

对思路中的7个方面分别进行相似度比较和打分，每个结点0-10分，对于每一点总结打分理由，然后输出最终得分。

>> Score the 7 aspects in the reasoning separately for similarity, 0–10 points for each item, summarize the reasons for each score, then output the final score.

若有诊疗思路2有明显信息遗漏，请大幅度扣分。一处遗漏扣3分，两处6分，依次递推，特别是明确的实体，如：药物类，病症类，活动类。

>> If Reasoning 2 has obvious missing information, deduct points significantly: 3 points for one omission, 6 points for two omissions, and so on, especially for concrete entities such as drugs, diseases, symptoms, and activities.

输出格式：

>> Output format:

【理由（精简回答）】：

>> [Reasons (concise answer)]:

【每点得分列表（仅输出列表，不要输出其他内容）】：[]

>> [Score list (only list, no other content)]: []

\end{promptbox}

\section{Data Demos}
\label{app:data-demo}

\subsection{CDRD}
\label{app:CDRD-demo}
\begin{promptbox}{结构化的临床诊断推理数据 (CDRD)}

\textbf{主症状} \\
>> \textbf{Chief Symptom}

咳嗽 \\
>> Cough

\hrulefill

\textbf{病情收集} \\
>> \textbf{Symptom Collection}%
\vspace{-0.5\baselineskip}
\begin{enumerate}[leftmargin=*]
    \item \textbf{咳嗽的性质}: 干咳、有痰 \\
    >> 1. \textbf{Nature of Cough}: Dry cough, productive cough (with sputum)
    \item \textbf{咳嗽的时间}: 几天、几个月、几年加重几天 \\
    >> 2. \textbf{Duration of Cough}: Days, months, years, aggravation for days
    \item \textbf{咳嗽的音色}: 咳嗽嘶哑、鸡鸣样咳嗽、咳嗽声音低微 \\
    >> 3. \textbf{Timbre of Cough}: Hoarse cough, whooping cough, weak cough sound
    \item \textbf{痰的性质}: 粘液性痰、泡沫样痰、脓性痰、痰中带血 \\
    >> 4. \textbf{Nature of Sputum}: Mucous sputum, frothy sputum, purulent sputum, blood-stained sputum
    \item \textbf{痰量}: 少、多 \\
    >> 5. \textbf{Sputum Amount}: Scant, copious
    \item \textbf{伴随症状}: 伴发热、伴胸痛、伴咳血、伴脓痰、伴哮喘、伴呼吸困难、咽干，咽痒，咽痛，鼻塞，喷嚏，流清涕 \\
    >> 6. \textbf{Accompanying Symptoms}: With fever, with chest pain, with hemoptysis, with purulent sputum, with asthma, with dyspnea, dry throat, itchy throat, sore throat, nasal congestion, sneezing, clear runny nose
\end{enumerate}

\hrulefill

\textbf{诊断考虑} \\
>> \textbf{Diagnosis Considerations}

\textbf{1. 上呼吸道感染} \\
>> \textbf{   1. Upper Respiratory Tract Infection}
\begin{itemize}[leftmargin=*]
    \item[] \textbf{疾病特点}: 起病较急，几天。表现为咳嗽，咽干，咽痒，甚至咽痛，伴有鼻部症状如鼻塞，喷嚏，流清涕。病原体多为病毒，少数为细菌。\\
    >> \textbf{Disease Features}: Acute onset, days. Manifested as cough, dry throat, itchy throat, even sore throat, accompanied by nasal symptoms such as nasal congestion, sneezing, clear runny nose. Pathogens are mostly viruses, occasionally bacteria.
    \item[] \textbf{实验室检查}: 白细胞计数正常或偏低，伴淋巴细胞比例偏高。\\
    >> \textbf{Laboratory Tests}: WBC count normal or low, with elevated lymphocyte percentage.
    \item[] \textbf{治疗反应}: 多饮水、保持室内空气流通和防治继发性细菌感染。一般5～7天痊愈，伴发并发症者可致病程迁延。\\
    >> \textbf{Treatment Response}: Drink plenty of water, maintain indoor air circulation, and prevent secondary bacterial infection. Generally recovers in 5-7 days, complications may prolong the course.
\end{itemize}

\textbf{2. 急性支气管炎} \\
>> \textbf{  2. Acute Bronchitis}
\begin{itemize}[leftmargin=*]
    \item[] \textbf{疾病特点}: 常起病较急，几天，发生于寒冷季节或气候突变时，也可由急性上呼吸道感染迁延不愈所致。全身症状较轻，可有发热。初为干咳或少量黏痰。\\
    >> \textbf{Disease Features}: Often acute onset, days, occurring in cold seasons or during weather changes, or may result from unresolved acute upper respiratory infection. Mild systemic symptoms, possible fever. Initially dry cough or scant mucous sputum.
    \item[] \textbf{实验室检查}: 周围血白细胞计数可正常，但由细菌感染引起者，可伴白细胞总数和中性粒细胞百分比升高。血沉加快，痰培养可见致病菌。\\
    >> \textbf{Laboratory Tests}: Peripheral WBC count may be normal, but in bacterial infection, total WBC and neutrophil percentage may be elevated. ESR increased, sputum culture may show pathogenic bacteria.
    \item[] \textbf{影像学检查}: X线胸片大多为肺纹理增强，少数无异常发现。\\
    >> \textbf{Imaging}: Chest X-ray mostly shows increased lung markings, few show no abnormalities.
\end{itemize}

\textbf{3. 肺部感染性疾病} \\
>> \textbf{   3. Pulmonary Infectious Diseases}
\begin{itemize}[leftmargin=*]
    \item[] \textbf{疾病特点}: 常见症状为咳嗽、咳痰，或原有呼吸道症状加重，并出现脓性痰或血痰，伴或不伴胸痛。病变范围大者可有呼吸困难、呼吸窘迫。大多数病人有发热。\\
    >> \textbf{Disease Features}: Common symptoms include cough, sputum production, or worsening of existing respiratory symptoms, with purulent or bloody sputum, with or without chest pain. Extensive lesions may cause dyspnea or respiratory distress. Most patients have fever.
    \item[] \textbf{确定病原体}: 血培养和痰培养分离到相同细菌，可确定为肺炎的病原菌。测定特异性IgM抗体滴度，可判断支原体、衣原体、嗜肺军团菌和病毒感染等。\\
    >> \textbf{Pathogen Identification}: Isolation of the same bacteria from blood and sputum cultures can confirm the pathogen of pneumonia. Measuring specific IgM antibody titers can identify Mycoplasma, Chlamydia, Legionella pneumophila, and viral infections.
    \item[] \textbf{影像学检查}: X线胸片多为肺实质浸润表现。\\
    >> \textbf{Imaging}: Chest X-ray mostly shows pulmonary parenchymal infiltration.
    \item[] \textbf{治疗反应}: 抗感染治疗：选择可能覆盖病原体的抗菌药物，抗感染治疗有效。\\
    >> \textbf{Treatment Response}: Anti-infective therapy: select antibiotics that may cover the pathogens, anti-infective treatment is effective.
\end{itemize}

\end{promptbox}

\vspace{0.5cm}

\subsection{Inquiry Example with Reasoning}
\label{app:inquiry-data-demo}
\begin{promptbox}{带推理的对话示例 (Inquiry Example with Six Reasoning Steps)}
\textbf{输入/提示词} \\
>> \textbf{Input/Prompt}

\textbf{历史:} \\
>> \textbf{History:}
\begin{itemize}[leftmargin=*]
    \item[] \textbf{患者}: 咳嗽、喉咙痒，术后伤口长肉芽发炎\\
    >> \textbf{Patient}: Cough, itchy throat, granulation tissue inflammation in postoperative wound.
    \item[] \textbf{医生}: 请问咳嗽持续了几天？是否有痰液或发热？\\ \textit{建议：多饮温水、保持空气湿润，术后伤口需每日消毒（如碘伏），若伤口红肿加重或出现脓液，请立即就医。}\\
    >> \textbf{Doctor}: How long has the cough lasted? Any sputum or fever? \\ \textit{>> Advice: Drink warm water, keep air humid, disinfect wound daily with iodine, if redness worsens or pus appears, seek medical help immediately.}
    \item[] \textbf{患者}: 咳嗽三四天了，没痰也不发烧。\\
    >> \textbf{Patient}: Coughing for 3-4 days, no sputum, no fever.
    \item[] \textbf{医生}: 请问咳嗽时是否伴有咽干、咽痒或鼻塞？术后伤口是否有胸痛或呼吸困难？\\ \textit{建议：多饮温水、保持空气湿润，可尝试使用右美沙芬止咳糖浆（具体用法遵说明书）。若咳嗽持续加重、出现痰液或发热，请及时就医。}\\
    >> \textbf{Doctor}: Is the cough accompanied by dry/itchy throat or nasal congestion? Any chest pain or difficulty breathing at the wound site? \\ \textit{>> Advice: Drink warm water, keep air humid, try Dextromethorphan cough syrup (follow instructions). If cough worsens, sputum or fever appears, seek medical help.}
    \item[] \textbf{患者}: 有咽痒，伤口没胸痛或呼吸困难。\\
    >> \textbf{Patient}: Have itchy throat, no chest pain or breathing difficulty at the wound.
    \item[] \textbf{医生}: 请问咳嗽时是否伴有鼻塞或流清涕？近期是否接触过花粉、宠物等过敏原？\\ \textit{建议：继续使用右美沙芬止咳糖浆（具体用法遵说明书），保持空气湿润，避免刺激性气体。若咳嗽持续超过1周或出现发热、痰液，请及时就医。术后伤口需每日碘伏消毒，若红肿加重或有脓液，请立即就诊。}\\
    >> \textbf{Doctor}: Is the cough accompanied by nasal congestion or clear runny nose? Have you been in contact with allergens such as pollen or pets recently? \\ \textit{>> Advice: Continue using Dextromethorphan cough syrup (follow instructions), keep air humid, avoid irritating gases. If cough persists >1 week or fever/sputum appears, seek medical help. Disinfect wound daily with iodine, if redness worsens or pus appears, seek medical help immediately.}
\end{itemize}

\noindent\textbf{最新消息/患者:} 没鼻塞流涕，没接触过敏原。

\noindent >> \textbf{Latest Message - Patient:} No nasal congestion or runny nose, no contact with allergens.

\hrulefill

\noindent\textbf{模型输出/推理与回复} \\
>> \textbf{Model Output - Reasoning \& Response}

\begin{lstlisting}[breaklines=true, breakindent=0pt, breakatwhitespace=false, columns=fullflexible, basicstyle=\small\ttfamily, frame=none, xleftmargin=0pt, xrightmargin=0pt, resetmargins=true]
【已知信息】
>> [Known Information] 
咳嗽3-4天，干咳伴咽痒，无痰、无发热、无鼻塞流涕，术后伤口无胸痛或呼吸困难，未接触过敏原。
>> Cough for 3-4 days, dry cough with itchy throat, no sputum, no fever, no nasal congestion or runny nose, no chest pain or dyspnea at wound site, no allergen contact.

【待解决用户需求】
>> [User Needs to Address] 
明确咳嗽病因，排除感染或术后并发症，指导用药及伤口护理。
>> Identify the cause of cough, exclude infection or postoperative complications, provide medication and wound care guidance.

【已提供给用户信息】
>> [Information Provided to User] 
已排除过敏性鼻炎典型症状（鼻塞流涕），建议使用右美沙芬止咳，伤口需碘伏消毒。
>> Excluded typical allergic rhinitis symptoms (nasal congestion and runny nose), advised Dextromethorphan for cough, wound needs iodine disinfection.

【诊断】
>> [Diagnosis] 
急性支气管炎
>> Acute Bronchitis

【待收集信息】
>> [Information to Collect] 
咳嗽是否伴有胸痛或呼吸困难？术后伤口是否有红肿加重或脓液？
>> Is the cough accompanied by chest pain or difficulty breathing? Is there worsening redness/swelling or pus at the wound site?

【回复策略】
>> [Response Strategy] 
聚焦关键点（胸痛/呼吸困难提示肺部感染，伤口感染需急诊），同步强化居家护理建议。
>> Focus on key points (chest pain/dyspnea suggests lung infection; wound infection needs ER), reinforce home care advice.

【回复】
>> [Response] 
请问咳嗽时是否伴有胸痛或呼吸困难？术后伤口目前是否有红肿加重或脓液？
>> Does the cough come with chest pain or difficulty breathing? Is the postoperative wound currently showing increased redness, swelling, or pus?

建议继续使用右美沙芬止咳糖浆（具体用法遵说明书），保持空气湿润，多饮温水。术后伤口需每日碘伏消毒，若出现红肿加重、脓液或发热，请立即就医。
>> I recommend continuing Dextromethorphan cough syrup (follow instructions), keeping the air humid, and drinking warm water. Disinfect the wound daily with iodine; if redness worsens, pus appears, or fever develops, seek medical attention immediately.
\end{lstlisting}
\end{promptbox}


\subsection{QA Pair}
\label{app:qa-pair-demo}
\begin{promptbox}{QA数据示例 (QA Data Examples)}

\textbf{【问题1】}

>> \textbf{[Question 1]}

请简述在对某一呼吸系统症状进行问诊时，应如何全面评估患者的呼吸道分泌物及相关表现？

>> Please briefly describe how to comprehensively evaluate a patient's respiratory secretions and related manifestations when taking a history for a respiratory symptom.

\vspace{\baselineskip}
\textbf{【参考答案1】}

>> \textbf{[Reference Answer 1]}

患者无咳嗽，但若出现咳嗽，应注意其音色是否嘶哑或呈鸡鸣样，声音是否低微；同时应观察咳出物的性质，如是否为粘液性、泡沫样、脓性或痰中带血；痰量的多少也是重要的评估内容，需判断为少量或大量；此外，还需注意是否伴有其他相关症状，以帮助明确病因和诊断。

>> The patient has no cough, but if cough occurs, attention should be paid to whether the sound is hoarse or whooping, and whether the voice is weak. At the same time, the nature of the sputum should be observed, such as whether it is mucous, frothy, purulent, or blood-stained sputum. The amount of sputum is also an important aspect of the assessment, and it should be determined as scant or copious. In addition, attention should be paid to whether there are other related symptoms to help clarify the etiology and diagnosis.

\hrulefill

\textbf{【问题2】}

>> \textbf{[Question 2]}

请简述肺炎的常见临床表现，并说明在临床诊断中为何不能仅依赖``咳嗽''这一症状进行判断。同时，请指出实验室检查在肺炎诊疗中的作用。

>> Please briefly describe the common clinical manifestations of pneumonia, and explain why clinical diagnosis cannot rely solely on the symptom of ``cough''. Also, please indicate the role of laboratory tests in the diagnosis and treatment of pneumonia.

\vspace{\baselineskip}
\textbf{【参考答案2】}

>> \textbf{[Reference Answer 2]}

肺炎是一种常见的呼吸系统感染性疾病，其典型临床表现包括咳嗽、咳痰，部分患者痰中可带血，常伴有发热、寒战和乏力等症状。在病情较重时，患者可能出现胸痛或呼吸困难。值得注意的是，并非所有患者都会出现咳嗽症状，因此在临床诊断中应综合考虑其他表现。实验室检查是明确诊断的重要手段，有助于判断感染类型及严重程度，从而指导后续治疗。

>> Pneumonia is a common respiratory infectious disease. Its typical clinical manifestations include cough and sputum production, some patients may have blood-stained sputum, often accompanied by fever, chills, and fatigue. In severe cases, patients may experience chest pain or dyspnea. It is worth noting that not all patients will present with cough symptoms, so clinical diagnosis should comprehensively consider other manifestations. Laboratory tests are an important means of confirming the diagnosis, helping to determine the type and severity of infection, thereby guiding subsequent treatment.

\end{promptbox}

\subsection{Patient Profile}
\label{app:patient-profile-demo}
\begin{promptbox}{患者画像示例 (Patient Profile Example)}

\textbf{年龄}: 45岁 \\
>> \textbf{Age}: 45 years old

\textbf{性别}: 男 \\
>> \textbf{Gender}: Male

\textbf{性格特征}: 较为顺从，对医生建议接受度高，但因病情反复而略显沮丧，经常询问病情进展。 \\
>> \textbf{Personality Traits}: Relatively compliant, high acceptance of doctor's advice, but slightly frustrated due to recurring illness, frequently asks about disease progression.

\textbf{语气特征}: 语气平和但带有担忧，对治疗效果和未来预后表示关心。 \\
>> \textbf{Tone Characteristics}: Peaceful tone but with worry, concerned about treatment effect and future prognosis.

\textbf{发言字数}: 每次发言字数中等，约25-45字。 \\
>> \textbf{Speaking Word Count}: Medium length per speech, about 25-45 words.

\textbf{症状}: 右肺中叶见磨玻璃样小结节，伴有长期干咳，夜间加重，有过敏性鼻炎史，偶有咽痛，吸烟。 \\
>> \textbf{Symptoms}: Ground-glass nodule in the right middle lobe, accompanied by chronic dry cough, worse at night, history of allergic rhinitis, occasional sore throat, smoker.

\textbf{诊断}: 咳嗽变异性哮喘，肺部磨玻璃结节（考虑良性）。 \\
>> \textbf{Diagnosis}: Cough variant asthma, pulmonary ground-glass nodule (considered benign).

\textbf{其他信息}: 使用吸入性糖皮质激素后症状减轻，但停药后易复发，有10年吸烟史，过敏性鼻炎定期发作。 \\
>> \textbf{Other Information}: Symptoms alleviated after using inhaled corticosteroids, but prone to relapse after stopping medication, 10-year smoking history, periodic attacks of allergic rhinitis.

\textbf{第一句话}: ``最近三个月一直干咳乏力。'' \\
>> \textbf{First Sentence}: ``I've had a dry cough and fatigue for the last three months.''

\end{promptbox}

\vspace{0.5cm}

\subsection{Real Clinical Inquiry Dialogue in Bench}
\label{app:real-inquiry-history-demo}
\begin{promptbox}{测评集中的真实临床问诊记录 (Real Clinical Inquiry Dialogue Example in Bench)}

这是你与患者的历史对话：

>> This is the history of your dialogue with the patient:

患者: 左右肩膀后背疼痛一个月，偶尔胸口疼，有高血压病史，未进行心电图或心脏相关检查。（匿名，女，35岁）

>> Patient: Pain in both shoulders and back for a month, occasional chest pain, history of hypertension, no ECG or heart-related examinations. (Anonymous, Female, 35 years old)

医生: 高血压控制得如何？最近有没有监测过血压？

>> Doctor: How is the hypertension controlled? Have you monitored your blood pressure recently?

医生: 平时有肥胖 血脂高吗

>> Doctor: Do you have obesity or high blood lipids?

患者: 胖，但是血脂高不高不知道，没查过，应该是不高

>> Patient: I am overweight, but I don't know if my blood lipids are high, I haven't checked, presumably not high.

医生: 不一定

>> Doctor: Not necessarily.

\vspace{\baselineskip}
这是患者的最新消息：

>> This is the latest message from the patient:

患者: 血压就高时候一百五左右，低压九十或者一百

>> Patient: When blood pressure is high, it is around 150, diastolic 90 or 100.

\end{promptbox}

\end{document}